\documentclass[10pt, a4paper, onecolumn]{article}

\usepackage{arxiv}

\usepackage[utf8]{inputenc} 
\usepackage[T1]{fontenc}    
\usepackage{hyperref}       
\usepackage{url}            
\usepackage{booktabs}       
\usepackage{amsfonts}       
\usepackage{nicefrac}       
\usepackage{microtype}      
\usepackage{lipsum}
\usepackage{amsmath}
\usepackage{amsfonts}
\usepackage{cite}
\usepackage{amssymb}
\usepackage{amsthm}
\usepackage{verbatim}
\usepackage{enumerate}
\usepackage{mdwlist} 
\usepackage{color}
\usepackage[utf8]{inputenc}

\usepackage{algpseudocode}
\usepackage{algorithmicx}
\usepackage{lineno}
\usepackage{framed,multirow}
\usepackage[final]{pdfpages}
\usepackage{latexsym}
\usepackage{wrapfig}

\usepackage{url}
\usepackage{xcolor}
\usepackage{graphicx, subfig}
\usepackage{verbatim}
\usepackage{bm}
\usepackage{authblk}

\title{Spatio-Spectral Networks for Color-Texture Analysis}

\author[1]{Leonardo F. S. Scabini}
\author[2]{Lucas C. Ribas}
\author[1,2]{Odemir M. Bruno}
\affil[1]{\small{S\~{a}o Carlos Institute of Physics, University of S\~{a}o Paulo (USP), PO Box 369, 13560-970, S\~{a}o Carlos, SP, Brazil. \protect\\Scientific Computing Group - http://scg.ifsc.usp.br}}
\affil[2]{\small{Institute of Mathematics and Computer Science, University of S\~{a}o Paulo (USP), USP, Avenida Trabalhador s\~ao-carlense, 400, 13566-590, S\~ao Carlos, SP, Brazil.}}

\begin{document}
\maketitle

\begin{abstract}
Texture is one of the most-studied visual attribute for image characterization since the 1960s. However, most hand-crafted descriptors are monochromatic, focusing on the gray scale images and discarding the color information. In this context, this work focus on a new method for color texture analysis considering all color channels in a more intrinsic approach. Our proposal consists of modeling color images as directed complex networks that we named Spatio-Spectral Network (SSN). Its topology includes within-channel edges that cover spatial patterns throughout individual image color channels, while between-channel edges tackle spectral properties of channel pairs in an opponent fashion. Image descriptors are obtained through a concise topological characterization of the modeled network in a multiscale approach with radially symmetric neighborhoods. Experiments with four datasets cover several aspects of color-texture analysis, and results demonstrate that SSN overcomes all the compared literature methods, including known deep convolutional networks, and also has the most stable performance between datasets, achieving $98.5(\pm1.1)$ of average accuracy against $97.1(\pm1.3)$ of MCND and $96.8(\pm3.2)$ of AlexNet. Additionally, an experiment verifies the performance of the methods under different color spaces, where results show that SSN also has higher performance and robustness.
\end{abstract}

\keywords{color-texture \and texture analysis \and feature extraction \and complex networks \and spatio-spectral networks}

\section{Introduction}

Texture is an abundant property in nature that allows to visually distinguish many things, and it is present not only in our common scale but also in macro and microscale, such as in satellite and microscopy imaging. There are various formal definitions to texture, for instance, according to Julesz \cite{julesz1962visual}, two texture are considered similar if their first and second order statistics are similar. We can also define texture in a simple way as a combination of local intensity constancy and/or variations that produce spatial patterns, roughly independently at different scales. Therefore, the challenge of texture analysis is to tackle these patterns in a multiscale manner, keeping a tradeoff between performance and computational complexity (cost). This has taken decades of study and heterogeneous literature, ranging from mathematical to bio-inspired methods.

Color images are present in the vast majority of current imaging devices, however, most of the texture analysis methods are monochromatic, i.e. they consider only one image channel, or the image luminance (grayscale). The true color information is usually lost during a grayscale conversion or processed separately with non-spatial approaches such as color statistics, histograms, etc. Convolutional networks have been employing the whole color information for various tasks such as object recognition \cite{krizhevsky2012imagenet,simonyan2014very,szegedy2016rethinking,he2016deep} and texture analysis \cite{cimpoi2016,bu2019deep}, with promising results. However, it is important to notice that even these methods do not consider the direct relation of pixels in different color channels. In fact, few works explore spatial patterns between color channels and their benefits for color-texture analysis.

The main contribution of this work is the proposal of a new method for color-texture analysis that performs a deep characterization of spatial patterns within-between color channels. This is achieved through a directed Spatio-Spectral Network (SSN) that models texture images creating connections pointing towards the gradient in a radially symmetric neighborhood, linking pixels from the same or different channels. A radius parameter defines a local window size, and each symmetric neighborhood contained in that region provides relevant color-texture information. The characterization is done through well-known network topological measures of low computational cost, such as the vertex degree and strength distributions. The combinations of different measures from the network structure provide a robust color-texture descriptor. The source code of the SSN method is available at GitHub \footnote{The script to compute SSN descriptors is available at \url{www.github.com/scabini/ssn}.}. We perform classification experiments to analyze the performance of our proposed descriptor and also compare our results with several methods from the literature in 4 color-texture datasets. Moreover, we analyze the robustness of each method for different color spaces (RGB, LAB, HSV and $I_1I_2I_3$) on each dataset.

\section{Theoretical Concepts and Review}\label{sec:TC}

\subsection{Texture Analysis}

The challenge of texture analysis lies in effectively characterizing local and global texture patterns while keeping a balance between performance and complexity (cost). Decades of research resulted in a wide range of methods \cite{liu2019bow,humeau2019texture}, where most of the techniques focus on grayscale images. Statistical methods explore measures based on grayscale co-occurrences in a local fashion, and the most diffused ones are the Gray Level Co-occurrence Matrices (GLCM) \cite{haralick} and the Local Binary Patterns (LBP) \cite{ojala1996comparative,ojala2002multiresolution}. These methods influenced future techniques, following the same principles. For instance, Local Phase Quantization (LPQ) \cite{ojansivu2008blur} computes local descriptors in a similar way to LBP but focusing on the problem of centrally symmetric blur. The Completed Local Binary Patterns (CLBP) \cite{guo2010completed} includes information not covered by the original LBP through the local difference sign-magnitude transform. Another approach to texture analysis consists of transforming the image into the frequency domain. There are various methods on this approach and most of them are based on the Gabor filters \cite{gabor2005} or Fourier spectrum \cite{fourier}. There is another category of texture analysis which explores image complexity and fits on the model-based paradigm, that includes methods based on fractal dimension \cite{fractal} and Complex Networks (CN) \cite{cantero2018importance,scabini2019multilayer}. The latter approach consists of modeling images as networks and using their topological properties for texture characterization.

\subsection{Color Vision and Color Texture}\label{sec:color}

Color is key information for the human visual system which helps to recognize things faster and to remember then better \cite{wichmann2002contributions}. The theory of opponent color processing \cite{foster1895text} determines that the human visual system combines the responses of different photoreceptor cells in an antagonistic manner. This happens because the wavelength of light for which the three types of cones (L, M, and S) respond overlap. Therefore, it is hypothesized that the interpretation process works in this way because for the visual system it is more efficient to store the differences between the responses of the cones rather than the response of each individual type of cone. The theory suggests that there are two opposing color channels: red versus green and blue versus yellow \cite{foster1895text}. In other words, according to this theory evolution has made the human visual system focus on variations between pairs of colors rather than individual colors as a way to improve color vision.

Computationally, the representation of colored images is given by different approaches, called color space. The most well-known color spaces can be classified into four main groups \cite{vandenbroucke2003color}, named by:

\begin{itemize}

\item Primary color spaces, based on the trichromatic retinal system. These methods assume that any color can be represented by the appropriate mixing of quantities of the three primary colors. The most common format is called RGB (Red, Green, and Blue). This format is widely used in many situations because most imaging devices work in this way. 


\item The luminance-chrominance color spaces, where one channel represents luminance and two channels describe the chrominance. A widely used format is known as LAB, or L*A*B*, or even CIELAB, and is a color space defined by the International Commission on Illumination (CIE) in 1976. This format is inspired by the theory of opponent color processing since channel L represents luminosity, channel A represents the red versus green component, and channel B represents the blue versus yellow component.

 
 \item Perceptual color space, where one of the best-known methods of this type is HSV (hue, saturation, and value) proposed in 1970. In this model the colors of each hue are arranged in a radial slice, that is, different from the three-dimensional space of the models mentioned above, the HSV format has a cylindrical geometry. In this space the hue represents the angular dimension, ranging from $0^{\circ}$ (red), $120^{\circ}$ (green) and $240^{\circ}$ (blue). The vertical axis represents $V$, which sets the brightness level from 0 (black) at the bottom to 1 (white) at the top. The distance from the center to the edge of the cylinder represents the value of $S$, which defines the saturation of the color, where the pure colors are at the maximum value of $S$, at the edges of the cylinder.
 
 
 \item The color spaces based on the concept of independent axes, which are obtained by statistical methods. An example of this type of method is the $I_1I_2I_3$ color space, which consists of computing components with the lowest possible correlation. Normally this color space is obtained through an RGB image, converted from a linear transformation. In this case, the channel $ I_1 $ represents the brightness ($I_1=\frac{R+G+B}{3}$), the channel $I_2$ the opponent between red and blue ($I_2=\frac{R-G}{2}$), and the channel $I_3$ the opponent between green, red and blue ($I_3=\frac{2G-R-B}{4}$).
 
\end{itemize}

Given an image represented by a certain color space, a method of color texture analysis should explore ways of characterizing the present spectral information. There are different approaches to color texture analysis in the literature and most of them are integrative, which separate color from texture. Integrative methods commonly compute traditional grayscale descriptors of each color channel, separately. In this case, any method of grayscale texture analysis can be applied by combining descriptors from each channel. Another type of approach is called pure color, which only considers the first order distribution of the color, not taking into account spatial aspects commonly analyzed in grayscale images. In this type of methods, the most widespread is based on color histograms \cite{hafner1995efficient}. This technique computes a compact summarization of the color distribution of the image. However, these methods do not consider the spatial interaction of pixels, as is done in texture analysis. Therefore, a common approach is to combine pure color descriptors with grayscale descriptors in parallel methods. In \cite{maenpaa2004,cernadas2017} the reader may consult an analysis between different integrative, pure color and parallel methods.

In some cases, integrative methods generate results similar to traditional grayscale methods. Authors have then questioned \cite{maenpaa2004} if the inclusion of color information is feasible for texture analysis, arguing that color and texture should be handled separately. However, it should be taken into account that few techniques consider all color information in a more intrinsic approach and that there are different concepts of color processing, even biological ones, that must be taken into account. For instance, the aforementioned opponent color processing theory of the human visual system is an interesting way of color characterization. Since texture consists of local patterns of intensity changes, color-texture information can be extracted by analyzing the spatial relationship between different colors. One of the first works approaching this idea \cite{rosenfeld1980multispectral} explores the interaction between pairs of color channels, and results indicate an increase in the textural information obtained, which is not available in the analysis of individual color channels. In a paper from 1998 \cite{opponentgabor1998multiscale} a technique based on filter banks is introduced for the characterization of colored texture. Monochromatic and opponent channels are obtained, calculated from the output of Gabor filters. More recently, some methods based on image complexity focused on the analysis of within-between channel aspects for color texture with fractal geometry \cite{casanova2016} and CNs \cite{sajunior2014,scabini2019multilayer}.

\subsection{Complex Networks}


The researches in CNs arise from the combination of graph theory, physics, and statistics, with the aim of analyzing large networks that derive from complex natural processes. Initially, works have shown that there are structural patterns in most of these networks, something that is not expected in a random network. This led to the definition of CN models that allow us to understand the structural properties of real networks. The most popular models are the scale-free \cite{scalefreeCN} and small-world \cite{smallworldCN} networks. Therefore, a new line of research has been opened for pattern recognition, where CNs are adopted as a tool for modeling and characterizing natural phenomena.


The concepts of CN are applied in several areas such as physics, biology, nanotechnology, neuroscience, sociology, among others \cite{aplicacoesRC}. Applying CNs to some problem consists of two main steps: i) modeling the problem as a network; ii) structural analysis of the resulting CN. The topological quantification of the CN allows us to arrive at important conclusions related to the system that it represents. For example, local vertex measurements can highlight important network regions, estimate their vulnerability, find groups of similar vertices, and etc.


A network can be defined mathematically by $N=\{V, E\}$, where $V=\{v_1, ..., v_n\}$ is a set of $n$ vertices and $E=\{a(v_i, v_j)\}$ a set of edges (or connections).
The edges can be weighted, representing a value that describes the weight of the connection between two vertices, or unweighted, indicating only if the connection exists. The edges can be either undirected, satisfying $a(v_i, v_j) = a(v_j, v_i)$, or directed, satisfying $\Diamond  (a(v_i, v_j) \neq a(v_j, v_i))$, i.e. $a(v_i, v_j)$ can be something other than $a(v_j, v_i)$.


The topology of a CN is defined by the patterns of its connections. To quantify it, measurements can be extracted for either individual vertices, vertex groups or globally for the entire network. One of the most commonly used measures is the vertex degree, which is the sum of its connections. Considering the sets $ V $ and $ E $, the degree $ k $ of each vertex $v_{i}$ can be calculated as follows:

\begin{equation}\label{eq:kin}
    k(v_{i}) = \sum_{\forall v_j \in V}\left\{
    \begin{array}{l}
    1 $, \ if  $a(v_i, v_j) \in E
    \\
    0 $,  \ otherwise $
    \end{array}
    \right
    .
    \end{equation}


Note that in this case the degree is calculated in a binary way since the sum considers only 1 if there is the edge, or 0 if it does not exist. If the network is weighted, the degree can also be weighted, metric commonly known by vertex strength. Therefore, the weight of all edges incident on the vertex is summed

\begin{equation}\label{eq:str}
s(v_{i}) = \sum_{\forall v_j \in V}\left\{
\begin{array}{l}
a(v_i, v_j) $, \ if $a(v_i, v_j) \in E
\\
0 $,  \ otherwise $
\end{array}
\right
.
\end{equation}


In directed networks it is possible to calculate the input and output degree of vertices according to the edge directions. The output degree $k(v_{i})_{out}$ represents the number of edges leaving $v_i$, and yields the same equation of the degree in undirected networks (Equation \ref{eq:kin}). To compute the input degree, it is necessary to invert the edge check

\begin{equation}\label{eq:kin}
k(v_{i})_{in} = \sum_{\forall v_j \in V}\left\{
\begin{array}{l}
1 $, \ if  $a(v_j, v_i) \in E
\\
0 $,  \ otherwise$
\end{array}
\right
.
\end{equation}
which then sums the number of edges pointing to $v_i$. Analogously we can compute the input and output strength of a vertex ($s(v_{i})_{in}$ and $s(v_{i})_{out}$) by summing the weight of its edges ($a$ instead of 1) according to its direction.









\subsection{Modeling of texture as CN}\label{sec:texturemodeling}

Since CNs are flexible structures that allow the analysis of several phenomena in the real world, it is possible to use them in image modeling. In this case, it becomes possible to transform a computer vision problem into a CN problem, which can be treated in different ways. The first step of this approach is the modeling of the network from the image, that is the definition of what are vertices and edges. This setting is variable and usually depends on the problem in question.
For texture modeling, the technique usually employed is to consider each pixel of the image as a vertex to build an undirected and weighted network. Consider a $ I $ gray image with $w * h$ pixels, with intensity levels $p(i)$ between $[0, L]$ ($L$ is the highest possible intensity value in the images). A network $N = \{V, E\}$ is obtained by constructing the set $V=\{v_1, ..., v_{w*h}\}$. The first work \cite{chalumeau2006} consider the absolute difference of intensity between pixels to define the weight of their connection. It is important to note that the intensity difference is not affected by changes in the average illumination of the image \cite{gonccalves2012walker}. Given a radius $r$ that defines a spatial boundary window, a new network $N^r$ is obtained where each vertex is connected to its neighbors with weight $\frac{|p(v_{i}) - p(v_{j})|}{L}$ (normalized absolute intensity difference) if $d(v_{i}, v_{j}) \leq r$, where $d(v_{i}, v_{j})$ represents the pixel Euclidean distance. This same modeling approach is also used by the methods proposed in \cite{wesley2015dynamic,gonccalves2016texture}. The connection weight inversely represent pixel similarity, where lower values means high similarity. However, this equation does not include spatial information on the connection weight, which led future works to propose new rules. In \cite{backes} the term $d$ was included on the calculation of the edge weight, giving equal importance to both the pixel intensity difference and its spatial position inside the connection neighborhood ($ \frac{1}{2r} (d(v_i, v_j) + r  \frac{|p(v_{i}) - p(v_{j})|}{L} )$). A different approach is introduced in \cite{scabini2015texture}, where intensity and distance are directly proportional ($\frac{|p(v_{i}) - p(v_{j})|}{L} \frac{d(v_i, v_j)}{r} $). The inclusion of the spatial information overcomes the limitation of previous methods where the connection weight towards pixels with the same intensity would be the same regardless of their distance to the central pixel.

The steps described so far results in a network $N^r$ with scale proportional to $r$, which limits the connection neighborhood. However, this is a regular network, as all vertices have the same number of connections (except for border vertices). Therefore, a transformation is needed in order to obtain a network with relevant topological information. In most works this is achieved through connection thresholding with an additional parameter $t$, therefore a new network $N^{r,t}$ is obtained by transforming its set of edges $E = \{a(v_{i}, v_{j}) | a(v_{i}, v_{j}) \leq t\}$. The resulting network then keep connected similar pixels, where $t$ controls the similarity level. It is intuitive to conclude that the resulting topology is directly influenced by the parameters $r$ and $t$. This allows a complete analysis of the network dynamic evolution from smaller to higher neighborhood sets and different levels of pixel similarity. The final texture characterization is then made through CN topological measures such as the vertex degree, strength, and others \cite{cnusp}.

The concepts of CN applied to texture analysis have been explored and improved in more recent works. In \cite{scabini2019multilayer} a new multilayer model is introduced for color texture analysis, where each network layer represents an image color channel, and its topology contains within-between channel connections in a spatial fashion. This work also proposes a new method for estimating optimal thresholding, and the use of the vertex clustering coefficient is also introduced for the network characterization, achieving promising results. In \cite{geovana2019classification} an interesting technique is proposed to build a vocabulary learned from CN properties and also to characterize the detected key points through CN, exploring the relevance of various topological measures.

\section{Spatio-Spectral Networks (SSN)}

We propose a new network modeling for color-texture characterization with various improvements over previous CN-based methods. Our method models the spatial relation of intra and inter-channel pixels through a directed CN, that we named Spatio-Spectral Network (SSN). Firstly, consider a color image $I$ with size $w*h*z$, i.e $w*h$ pixels with $z$ colors whose values range from $[0,L]$, and a network $N$ defined by a tuple $N=\{V_N,E_N\}$, where $V_N$ represents the network vertices and $E_N$ its edges. The vertex set is created as in \cite{scabini2019multilayer}, where each image pixel $i$ is mapped as a vertex for each color-channel, thus $V_N=\{v_1,...,v_{w*h*z}\}$. This creates a multilayer network, where each layer represents one image color-channel and each vertex $v_i$ carry a pair of coordinates $(x,y)$, indicating the position of the pixel that it represents, and a value $p(v_i) \in L$ indicating the pixel intensity value on its respective color-channel. 

In order to create the connections, previous works usually adopt a set of radii $R=\{r_1,...,r_n\}$ to limit the size of the vertex neighborhood, i.e. a set of sliding windows of radius $r_i$ defines a distance limit for vertices to be connected, resulting in a set of $N^{r_i}$ networks.
In other words, a vertex $v_i$ is connected to a pixel $v_j$ if it is inside its neighborhood $G_{v_i}^{r_i} = \{v_j \in V_N | d(v_i, v_j) \leq r_i\}$, where $d(v_i, v_j)$ is the 2-D Euclidean distance between the vertices. Therefore, as in \cite{scabini2019multilayer}, this neighborhood covers vertices in all color-channels because it considers only the spatial position of the pixels to compute its distance. This process allows to access the dynamic evolution as $r$ increases by analyzing each network $N^{r_1},...,N^{r_n}$, and have demonstrated to be effective for color texture characterization \cite{scabini2019multilayer}. On the other hand, it is possible to notice that $\{G_{v_i}^{r_1},..., G_{v_i}^{r_{i-1}}\} \subsetneqq   G_{v_i}^{r_{i}}$ for a set of increasing radii $R=\{1,2,...,r_n\}$, and $\forall (r_i,r_j)$, $ G_{v_i}^{r_i} \neq G_{v_i}^{r_j}$. This means that each neighborhood contains all neighborhoods from previous radius, which leads to redundancy between networks $N^{r_1},...,N^{r_n}$.

\begin{figure}[!htb]
    \centering
    \subfloat[Standard CN neighboring ($r=2$ and $r=3$)]{\includegraphics[width=0.2\linewidth]{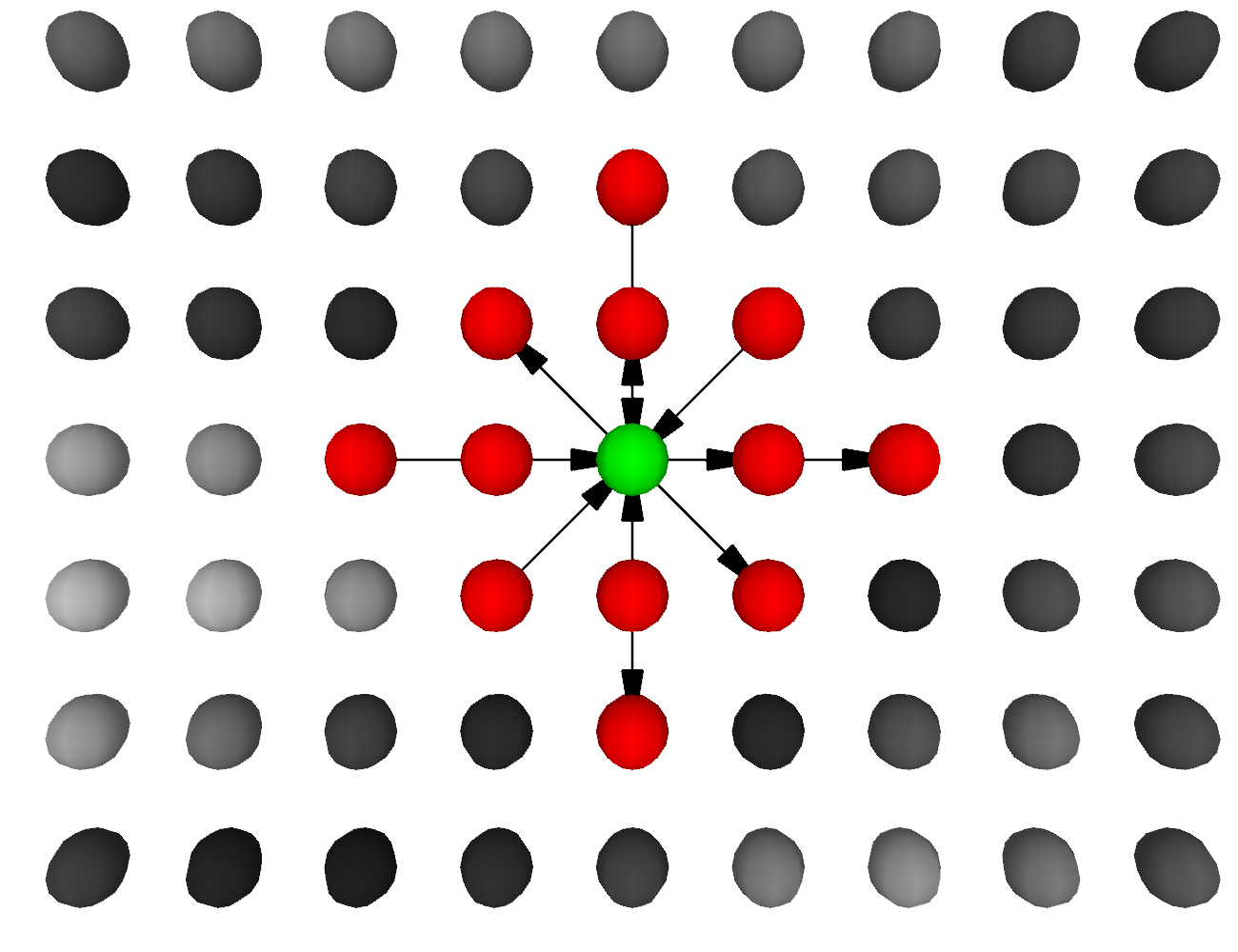} \ \ \ \ \ \ \ \ \ \ \includegraphics[width=0.2\linewidth]{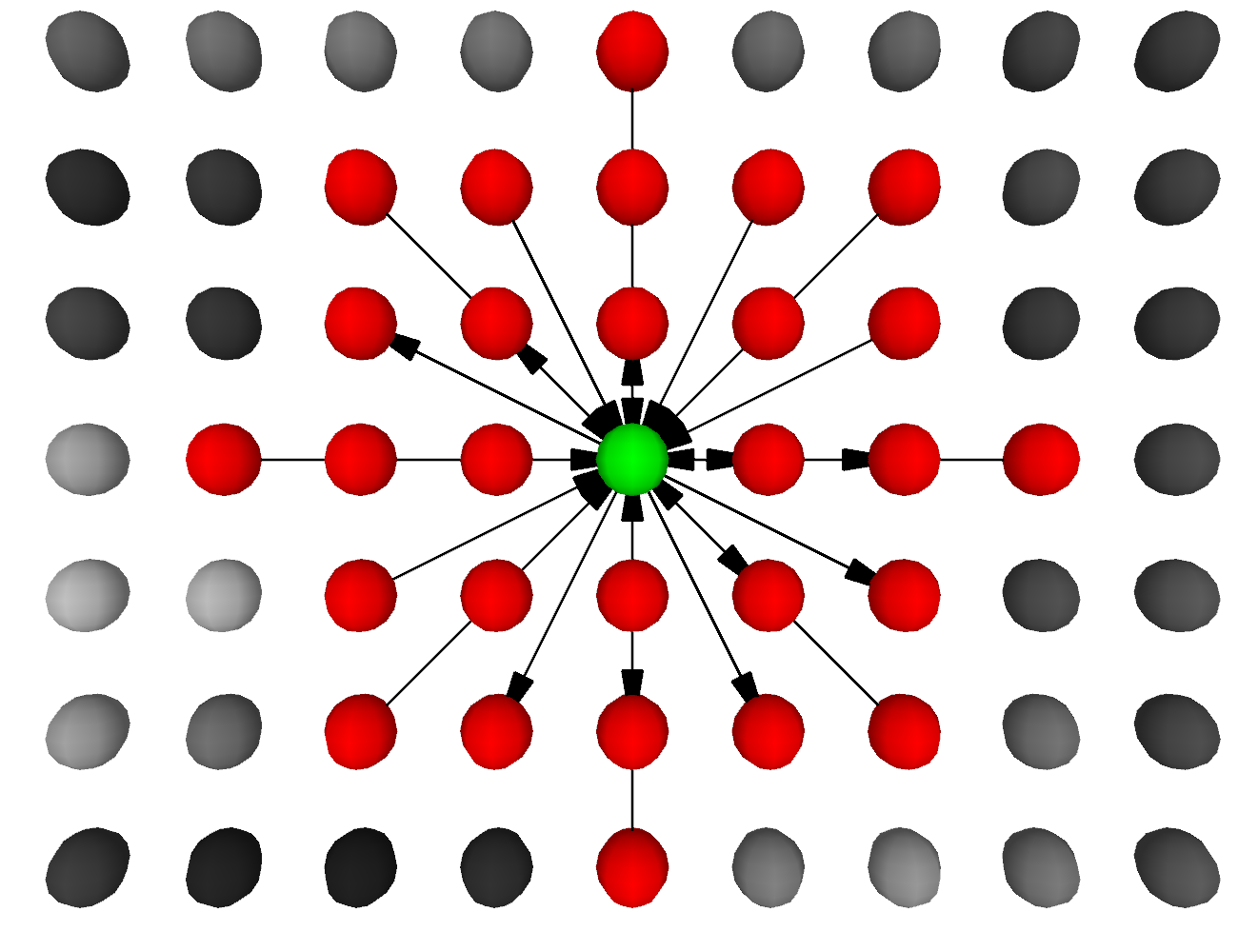}}\\ 
    \subfloat[Radially symmetric CN neighboring ($r=2$ and $r=3$)]{\includegraphics[width=0.2\linewidth]{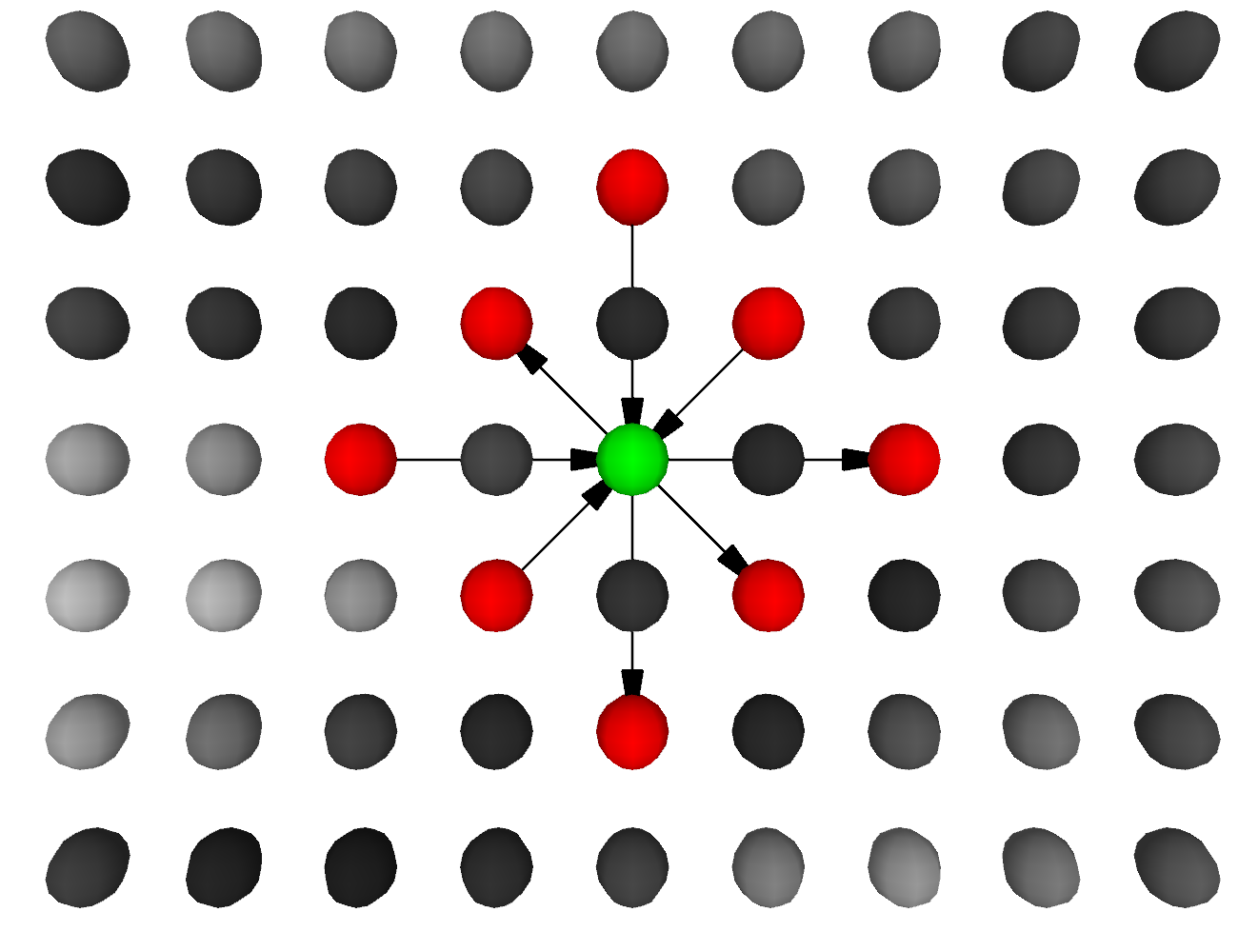}  \ \ \ \ \ \ \ \ \ \   \includegraphics[width=0.2\linewidth]{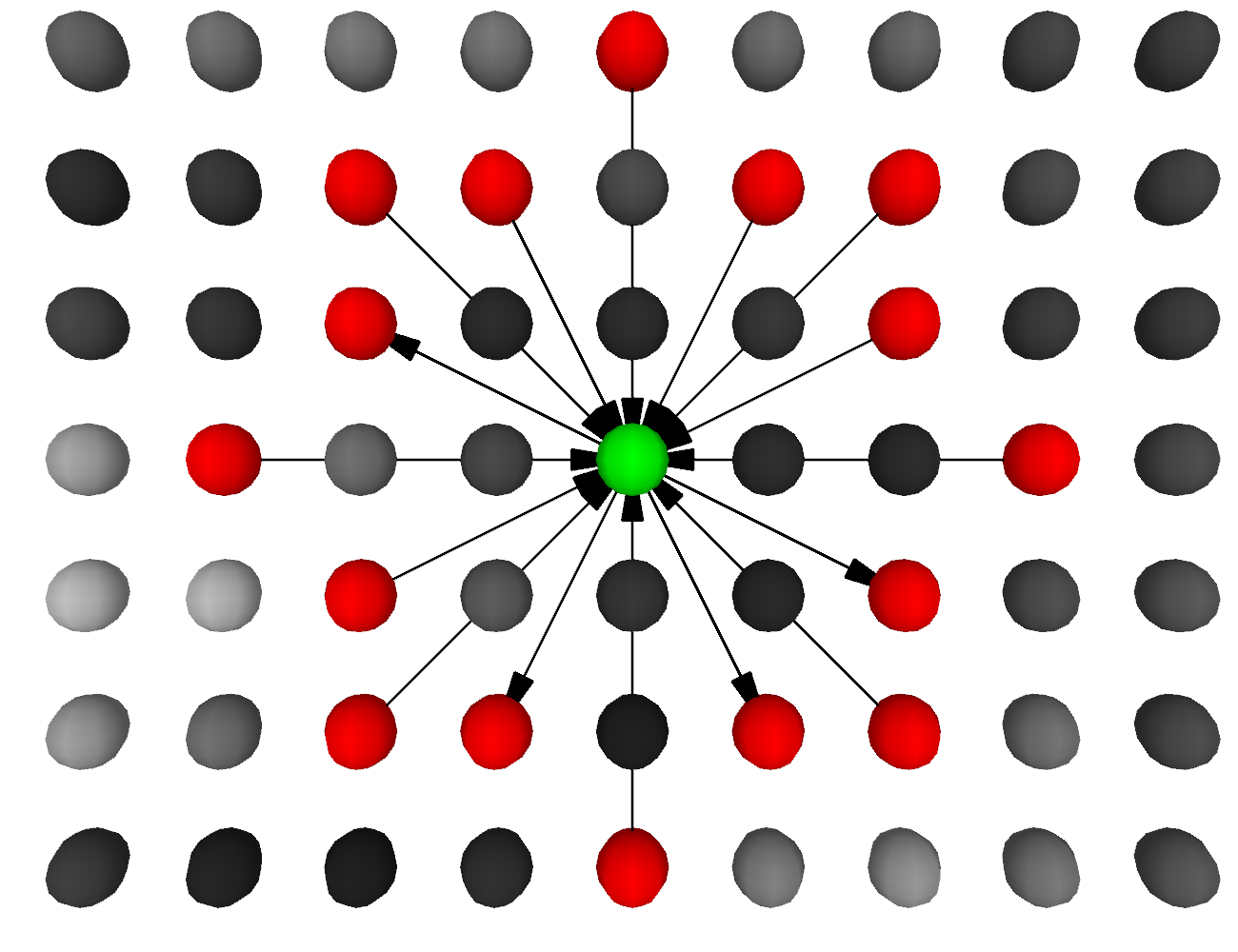}} \ \ \ \ \ \ \ \ 
    \subfloat[Proposed radially symmetric neighboring in a multilayer CN ($r=2$)]{\includegraphics[angle=90, width=0.35\linewidth]{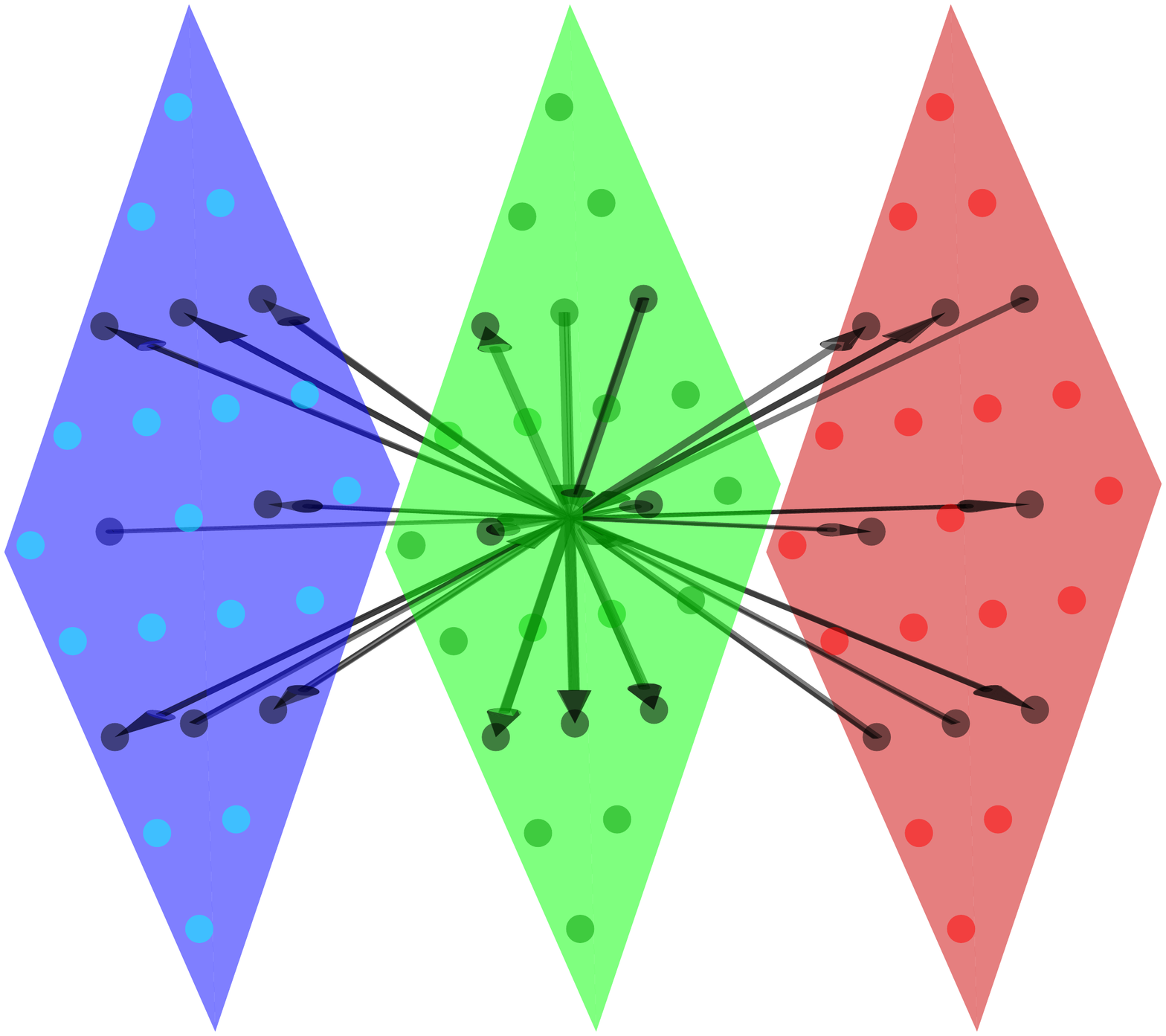}}
    
     \subfloat[$N^{r=1}$]{\includegraphics[width=0.25\linewidth]{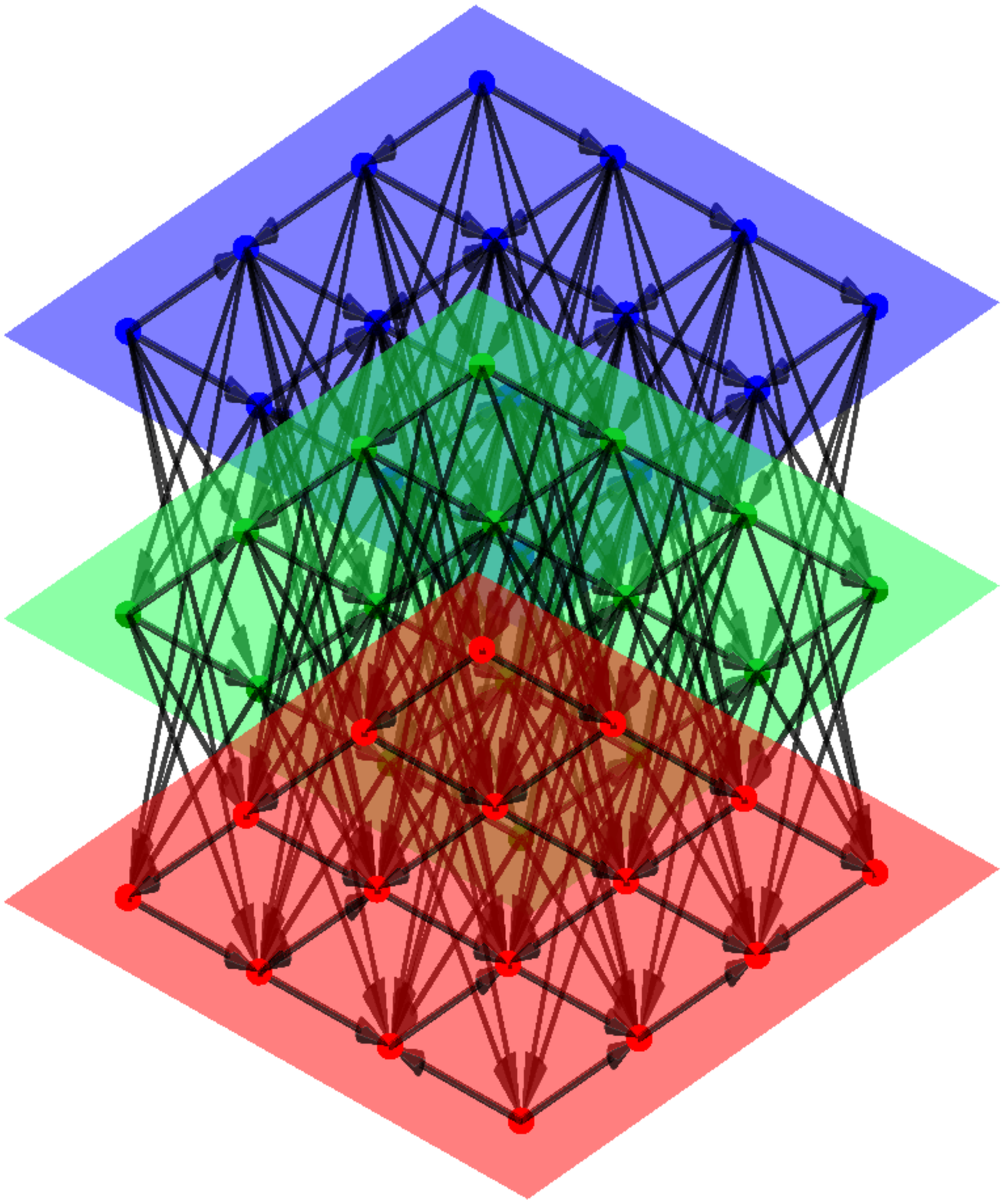}} \ \ \ \ \ \  \subfloat[$W^{r=1}$]{\includegraphics[width=0.25\linewidth]{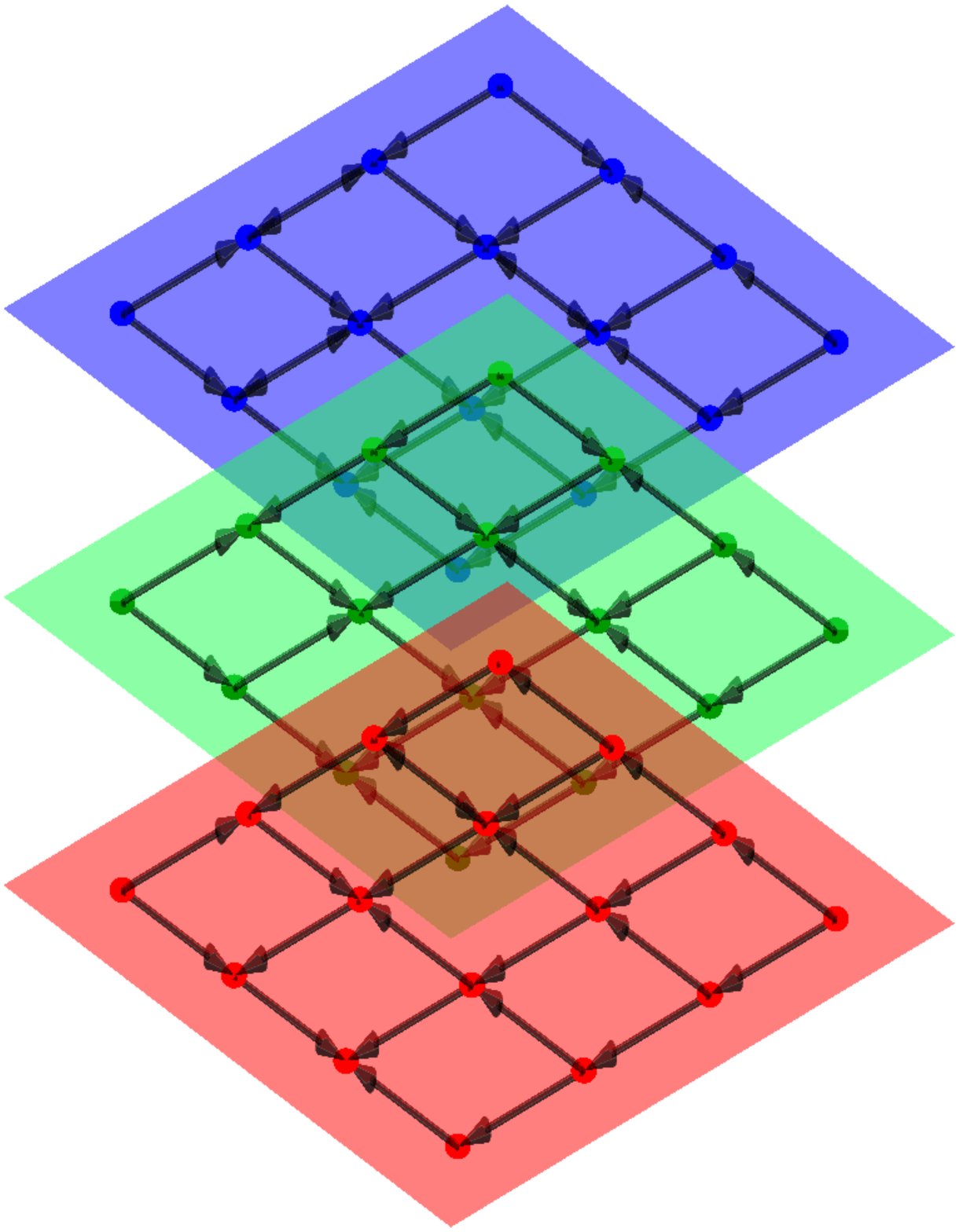}} \ \ \ \ \ \  \subfloat[$B^{r=1}$]{\includegraphics[width=0.25\linewidth]{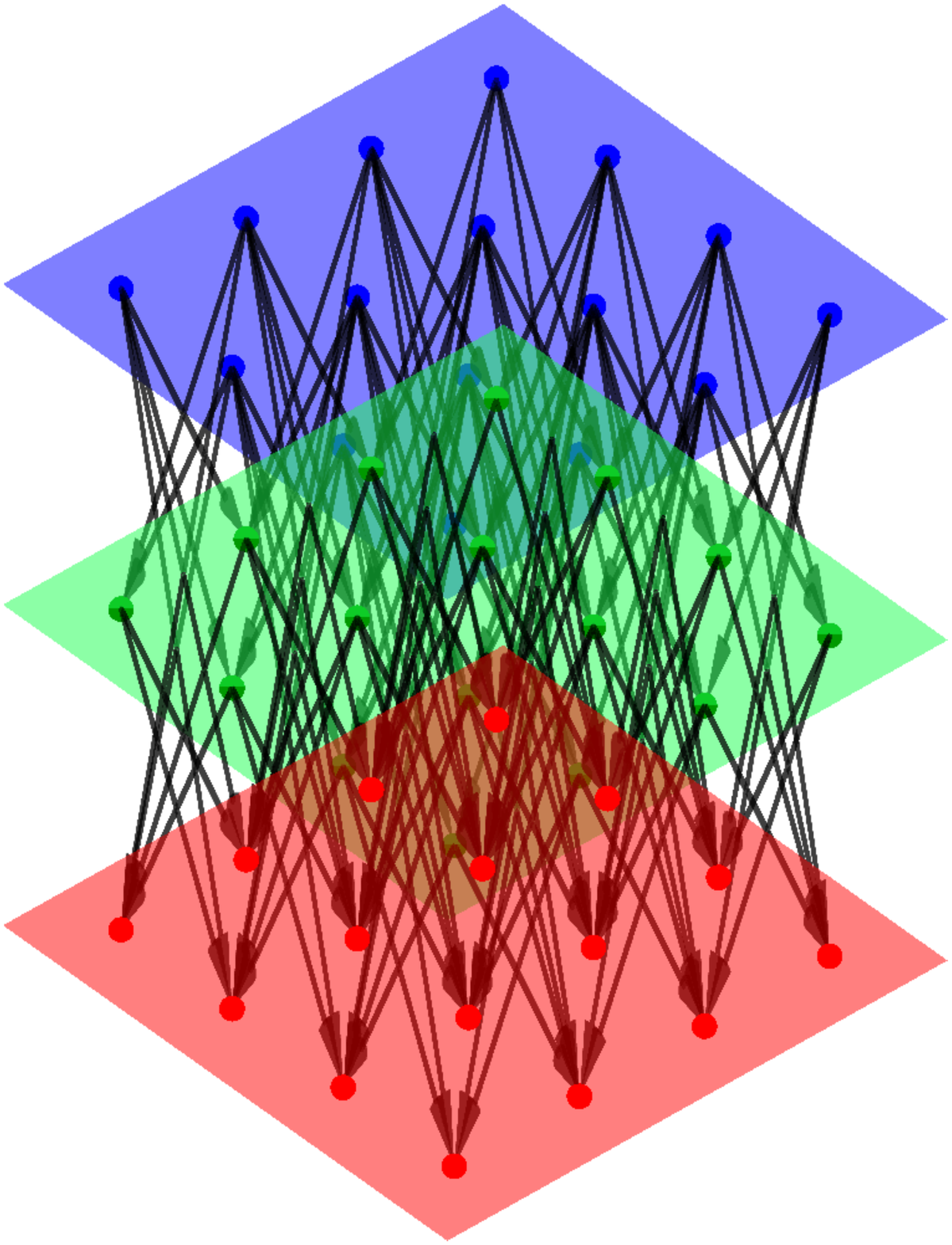}}
    
    \caption{\label{fig:neighborhood} Two neighboring approaches for CN connection creation in a single-channel image (a,b) and the extension of the symmetric neighborhood for a 3-channel image (c). Figures d, e and f shows the multilayer structure of the spatio-spectral networks modelled for $r=1$.}
\end{figure}

In this context, we propose a radially symmetric modeling criteria by redefining the neighborhood as $G_{v_i}^{r_i} = \{v_j \in V_N | r_{i-1} < d(v_i, v_j) \leq r_i\}$, i.e. it considers only vertices in a distance range between the considered and the previous radius, thus $\{G_{v_i}^{r_1},..., G_{v_i}^{r_{i-1}}\} \nsubseteq G_{v_i}^{r_{i}}$. It is important to notice that we include vertices with distance 0 in the neighborhood $G_{v_i}^{1}$, which is the case when a pixel connects to itself in different color-channels. The neighborhood definition can also be formulated as a combination of the distance $r_i$ and the number of neighbors $p_i$, which is usually adopted in LBP-based methods \cite{guo2010completed}. In our case, as the neighborhood covers all the color channels, we can represent the vertex neighbors for radius 1 by ($r_1=1$, $p_1=4z + z-1$), as the vertex connects to 4 neighbors in each channel plus itself on the other $z-1$ channels. Analogously, radius 2 creates a neighborhood ($r_2=2$, $p_2=8z + z-1$), and so on. To illustrate this concept, Figure \ref{fig:neighborhood} shows a CN modelled for 1 image channel and highlights the difference between the standard (a-b) and the radially symmetric (c-d) neighboring. In our proposal, the radially symmetric neighborhood is then extended for all image channels (e).

The definitions given so far concerns only the pixel neighborhood in which vertices will connect, therefore the next step is to define the connection creation. The weight of the connection $a(v_i, v_j)$ between pairs of vertices are defined by their absolute intensity difference directly proportional to their Euclidean distance on the image

\begin{equation}\label{eq:connectionWeight}
a(v_i, v_j) = \frac{(|p(v_i) - p(v_j)| + 1)(d+1)-1}{(L + 1)(r+1)-1}
\end{equation}

This is a modification over the equations proposed in \cite{scabini2015texture,scabini2019multilayer} so that neither side of the multiplication cancels the other when the intensity difference or the distance is 0 (same pixel in different channels), and to generate uniform values between $[0,1]$. It is important to notice that according to the proposed neighboring, the creation of connections from all to all channels with the given connection weight implies that the network somehow performs opponent color processing in a spatial fashion. Therefore, the SSN is a combination of CN and a bio-inspired approach based on the Opponent-Process Theory \cite{foster1895text}, but considers the opposing color pairs red versus green, red versus blue and green versus blue instead of red versus green and blue versus yellow. If another color space than RGB is used, then different opposing color pairs will be considered acording to that space.

As previously mentioned, most CN-based approaches employ weighted undirected networks, which needs thresholding techniques in order to obtain relevant topological information for analysis. This happens because a network that connects all pixels inside a fixed radius have constant topological measures such as degree, strength, clustering, etc. However, the thresholding leads to additional parameters that need to be tuned, which have been a drawback of previous works where authors explored costly techniques for optimal threshold selection \cite{scabini2015texture,scabini2019multilayer}. In this sense, we propose a directed technique that eliminates the need for thresholding, reducing the method parameters to only the set of radius $R=\{r_1,...,r_n\}$. This is achieved by associating the direction of the connection to the direction of the gradient, i.e. towards the pixel of higher intensity. This idea was first employed in previous work for grayscale texture characterization \cite{ribas2018fusion}, and here we extend its definitions for multilayer networks of color images, along with our new connection weight equation (Equation \ref{eq:connectionWeight}). Consider a network $N^{r_i}=\{V_N,E_N\}$, its set of edges is defined by

\begin{equation}
E_N = \left \{a(v_i,v_j) \in E \mid r_{i-1} < d(v_i,v_j) \leqslant r_i   \wedge  p(v_i) < p(v_j) \right \},
\end{equation}
and when $p(v_i) = p(v_j)$ the edge is bidirectional ($a(v_i,v_j) = a(v_j,v_i)$). This process generates a network that contains relevant topological information reflected on its directional connection patterns, which can be quantified through directed measures such as the input and output degree and strength of vertices, eliminating the need for connection cutting.

The spatio-spectral nature of our network emerges from the patterns of within-between channel connections which includes information of the image gradient through edge directions. To highlight this information, we divide the original network $N^{r_i}(V_N, E_N)$ as in \cite{scabini2019multilayer}, obtaining two additional networks, being the first $W^{r_i}(V_W, E_N)$ whose edges are a subset of $N$ that contains within-channel connections, thus $\forall w(v_i, v_j) \in E_N$, $E_W= a(v_i, v_j)| p(v_i, z) = p(v_j, z)$, where $p(v_i, z)$ return the channel/layer of $v_i$. The second network $B^{r_i}(V_B,E_B)$ represents between-channel connections, then $\forall a(v_i, v_j) \in E_N $, $ E_B = a(v_i, v_j) | p(v_i, z) \neq p(v_j, z)$. The vertex set of each network $N$, $W$ and $B$ are the same ($V_N = V_W = V_B$) as we only divide its edges. Figure \ref{fig:neighborhood} (d, e and f) illustrates the structure of the 3 networks for radius $r=1$. By quantifying their topology, it is possible to obtain rich color-texture information for image characterization, as we discuss in the following.



\subsection{Network characterization}

To characterize the directed SSN, we propose the use of traditional centrality measures computed for each vertex considering its edge directions, which is the input and output degree and strength (See equation \ref{eq:kin}). These measures can be effectively computed during the network modeling as only the vertex neighbors must be visited for the calculation, therefore there is no significant additional cost for the network characterization. Our approach cost is then smaller than the method of \cite{scabini2019multilayer}. This one uses the vertex clustering coefficient, which needs to store the network to visit the neighbors of the neighbors of each vertex.

\begin{figure}[!htb]
    \centering
    \subfloat[Input]{\includegraphics[width=0.11\linewidth]{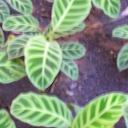}}\\
    \subfloat[$k_{in}$]{\includegraphics[width=0.6\linewidth]{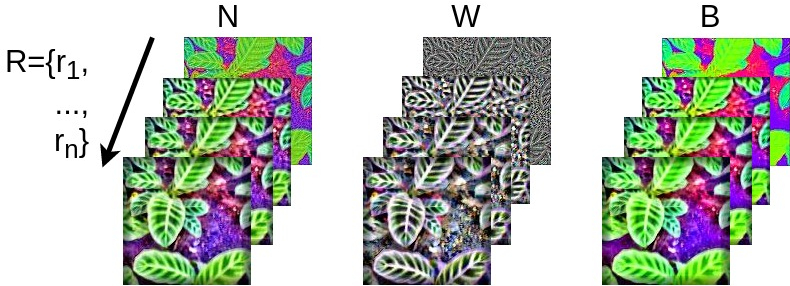}}\\
    \subfloat[$s_{in}$]{\includegraphics[width=0.6\linewidth]{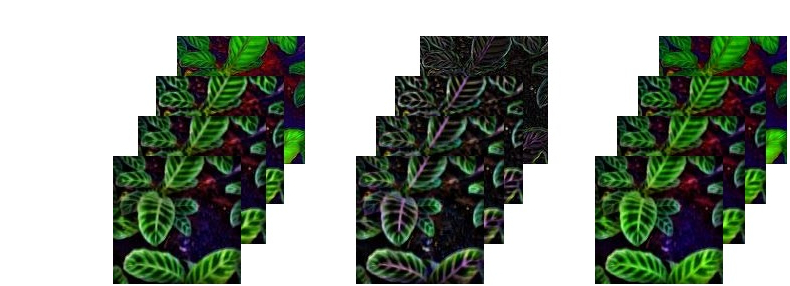}}\\
    \subfloat[$s_{out}$]{\includegraphics[width=0.6\linewidth]{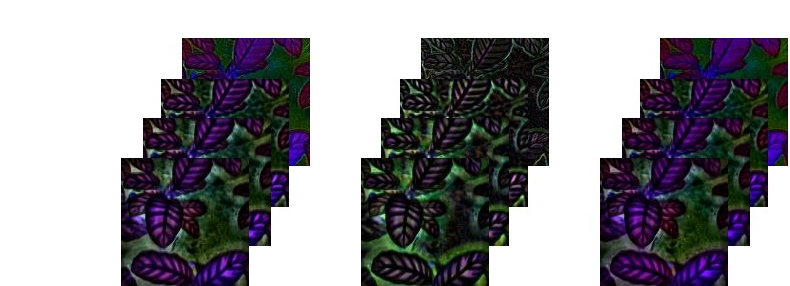}}\\
    
    \caption{\label{fig:visualexp2} Feature maps for three directed centrality measures (b-d) obtained from SSN modeled for the input color-texture (a) with a set of radii $R$. A color image is obtained by converting vertex measures in each network layer into an intensity value to its corresponding color-channel.}
\end{figure}

Notice that $\forall v_i \in V$, $k(v_{i})_{in} + k(v_{i})_{out} = k_{max}^r$, where $k_{max}^r$ indicates the maximum number of possible connections for $r$. This means that the input and output degree are a linear transformation of each other in the function of the network max degree, therefore we use only the input degree. It is not possible to make the same assumption for the vertex strength, as it also depends on the distribution of image pixel values in the connection neighborhood, thus we use both the input and output strength. In this context, the network characterization is performed combining the three measures (input degree, input strength, and output strength). Each topological measure highlights different texture patterns of the modeled image, as we show in Figure \ref{fig:visualexp2} for each network. We consider a set of feature maps all the topological information obtained through the characterization of the networks $N$, $W$ and $B$ with the aforementioned 3 centrality measures ($k_{in}$, $s_{in}$ and $k_{out}$) in a multiscale fashion with a set of increasing radius $R=\{1,...,r_n\}$ given $r_n$. In the qualitative analysis shown in Figure \ref{fig:visualexp2}, it is possible to notice how the feature maps highlight a wide range of color-texture patterns.

\begin{figure}[!htb]
    \centering
    \subfloat[Input textures]{\includegraphics[width=0.12\linewidth]{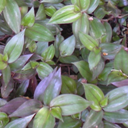} \ \ \ \ \ \includegraphics[width=0.12\linewidth]{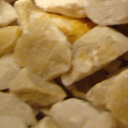}}\\
    \subfloat[Distribution of $k_{in}$ for each layer of $N^{r=4}$.]{\includegraphics[width=0.23\linewidth]{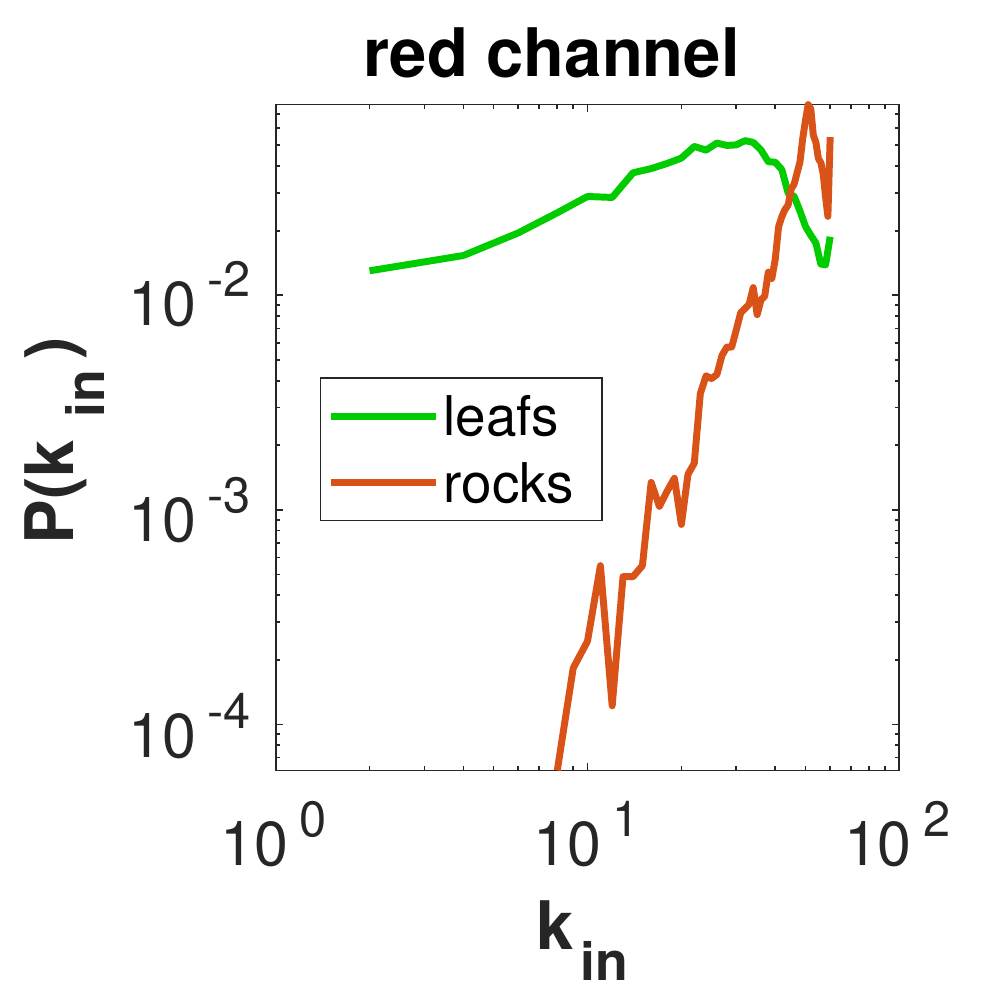} \includegraphics[width=0.23\linewidth]{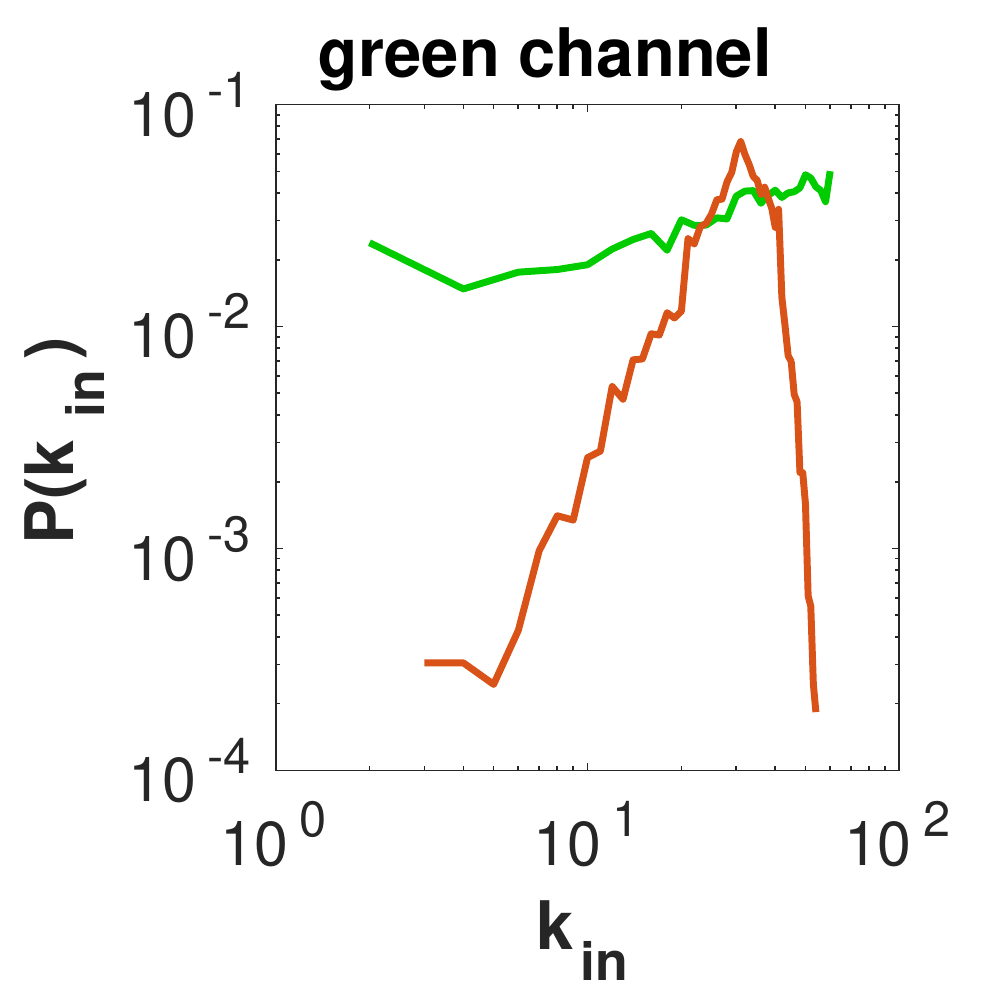} \includegraphics[width=0.23\linewidth]{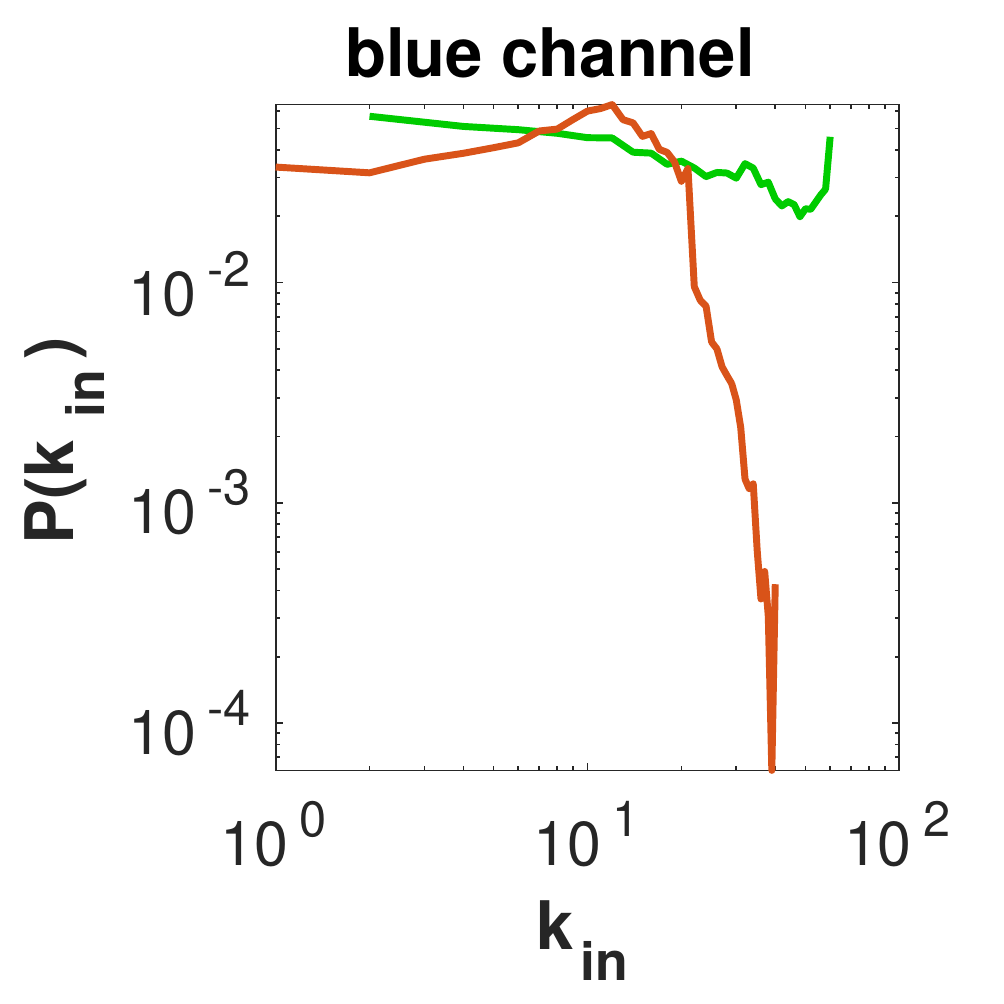}} \\
    
    \subfloat[Distribution of $s_{in}$ for each layer of $N^{r=4}$.]{\includegraphics[width=0.23\linewidth]{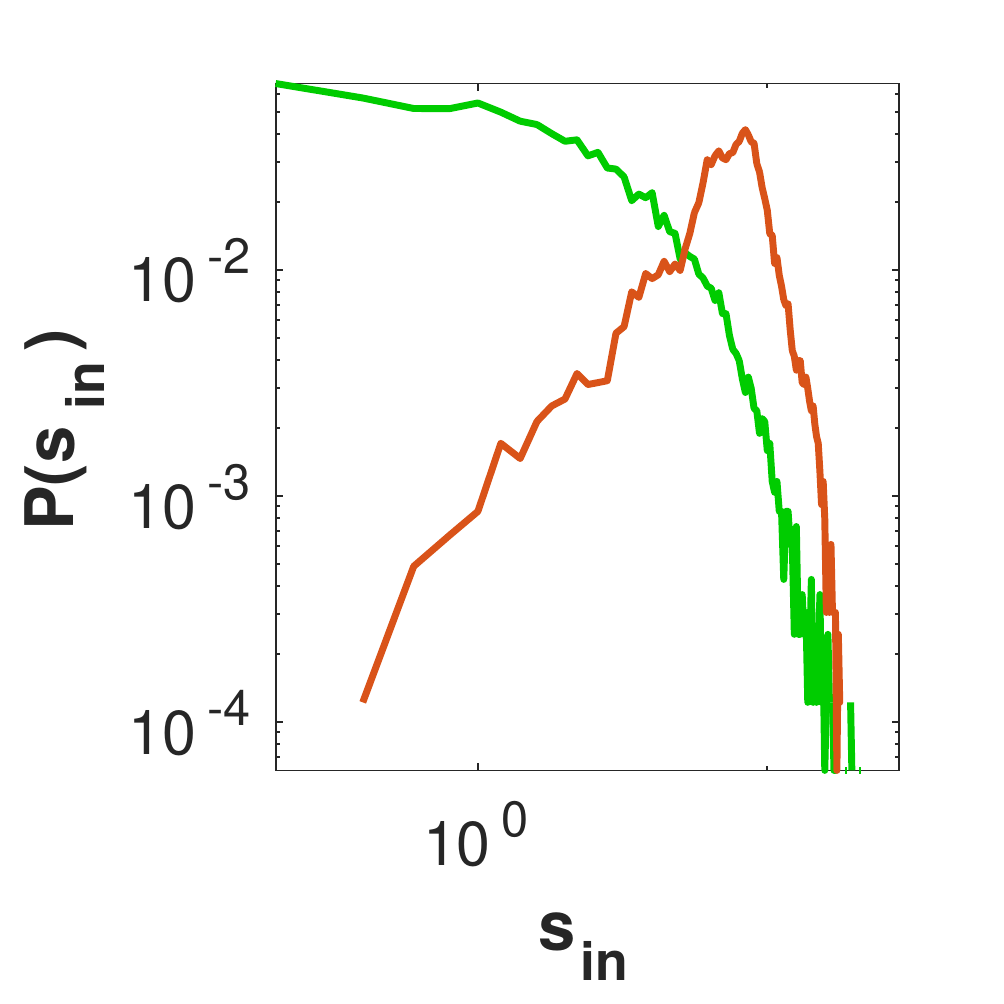} \includegraphics[width=0.23\linewidth]{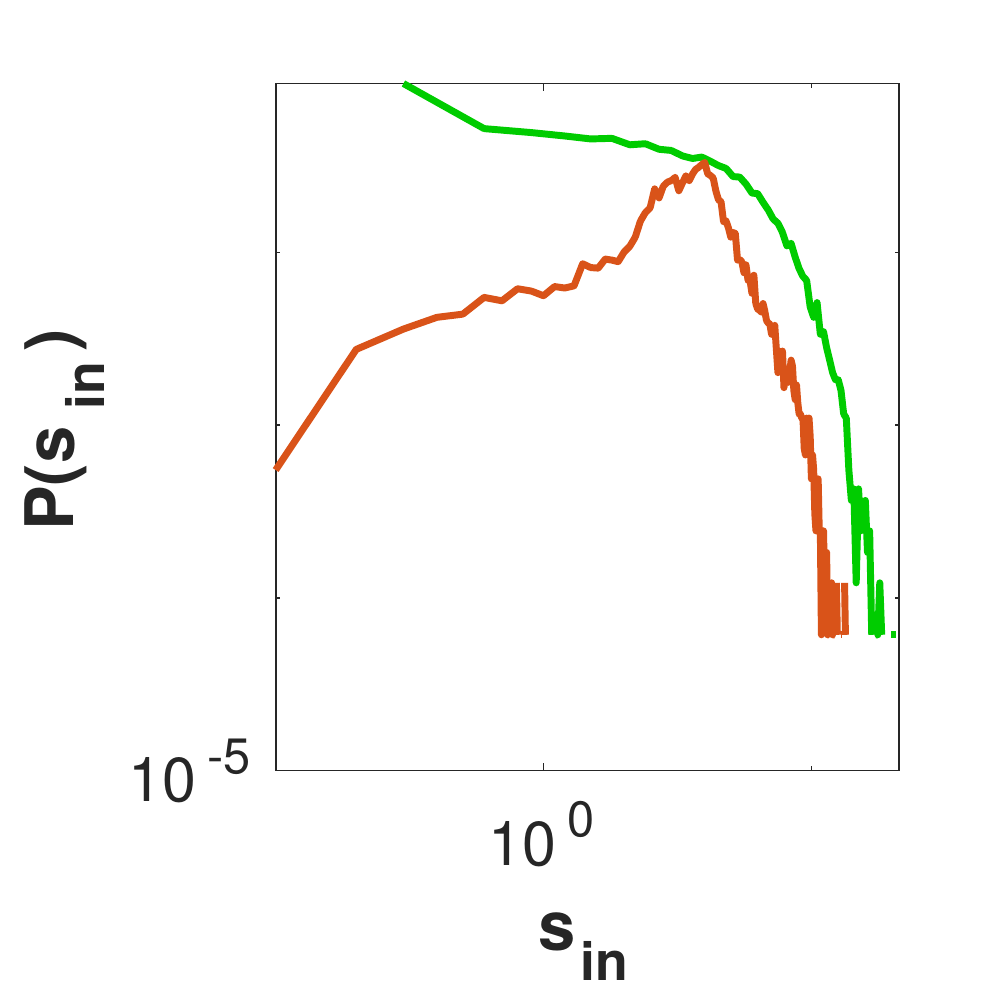} \includegraphics[width=0.23\linewidth]{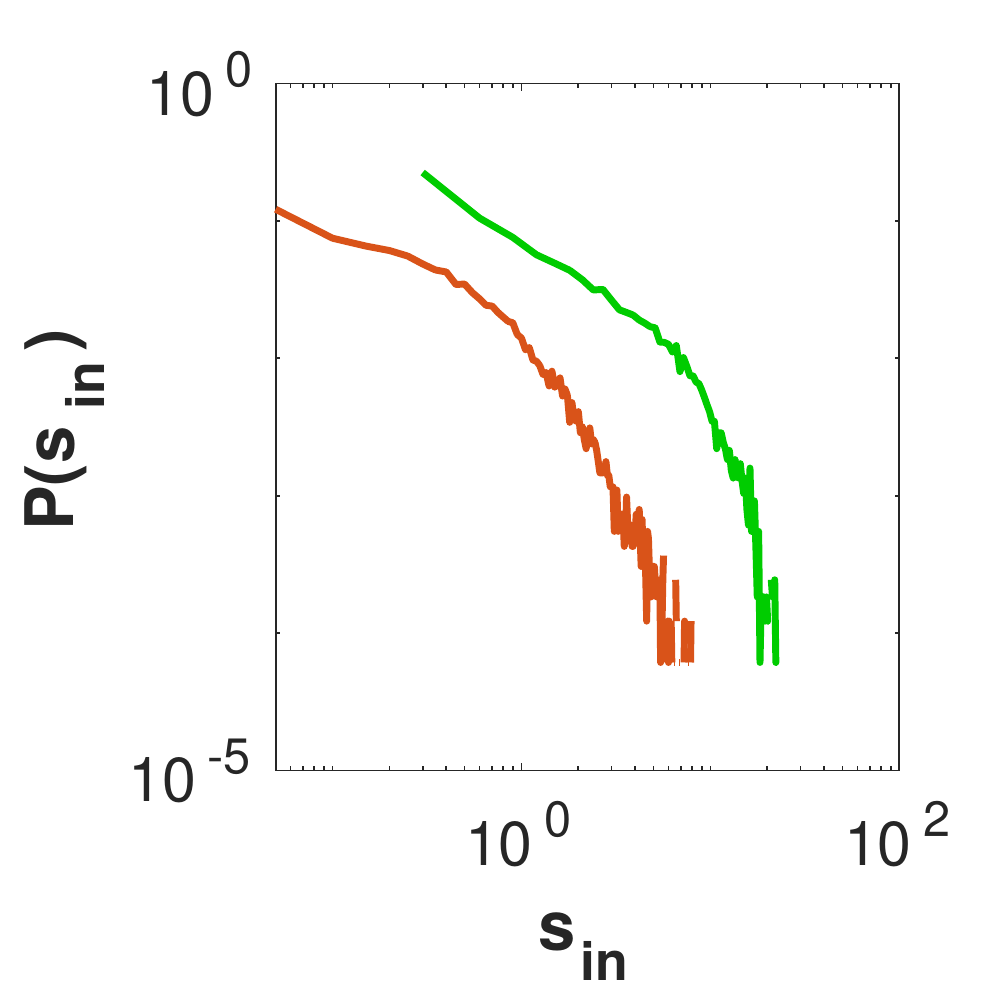}}
    
    \subfloat[Distribution of $s_{out}$ for each layer of $N^{r=4}$.]{\includegraphics[width=0.23\linewidth]{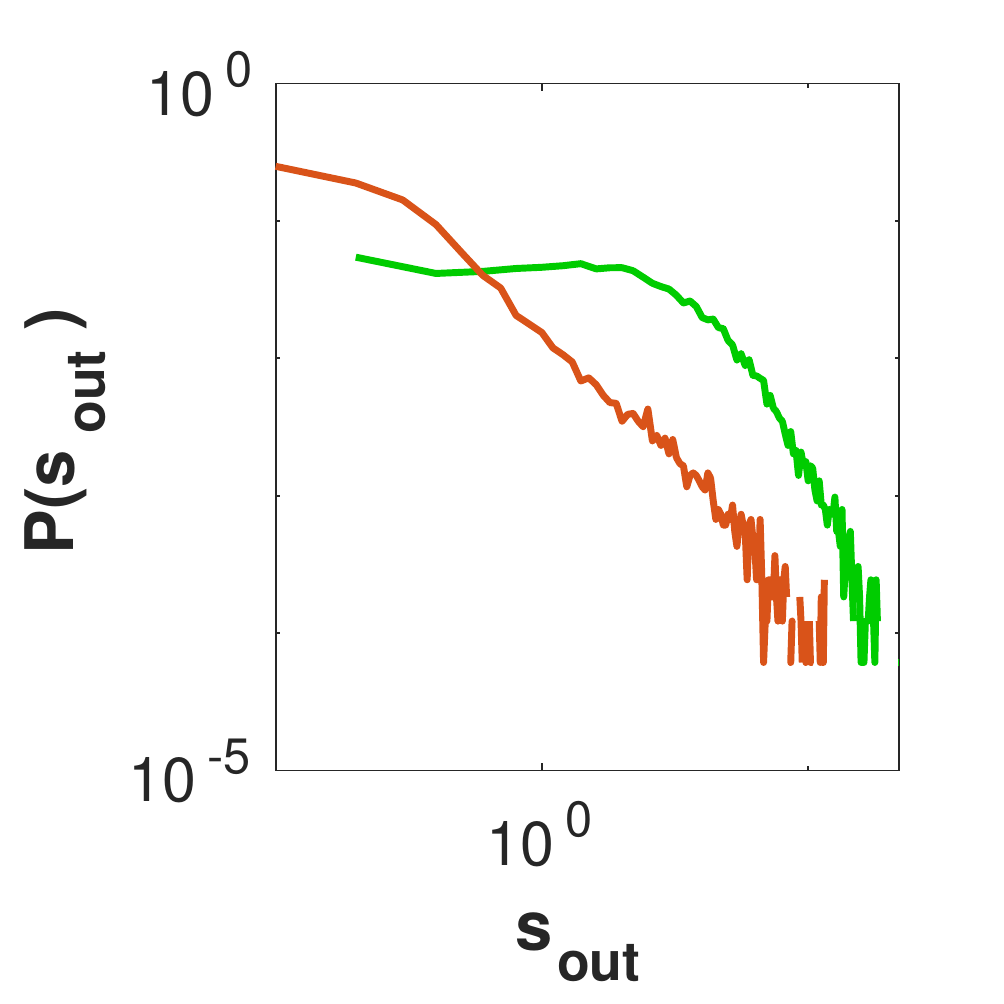} \includegraphics[width=0.23\linewidth]{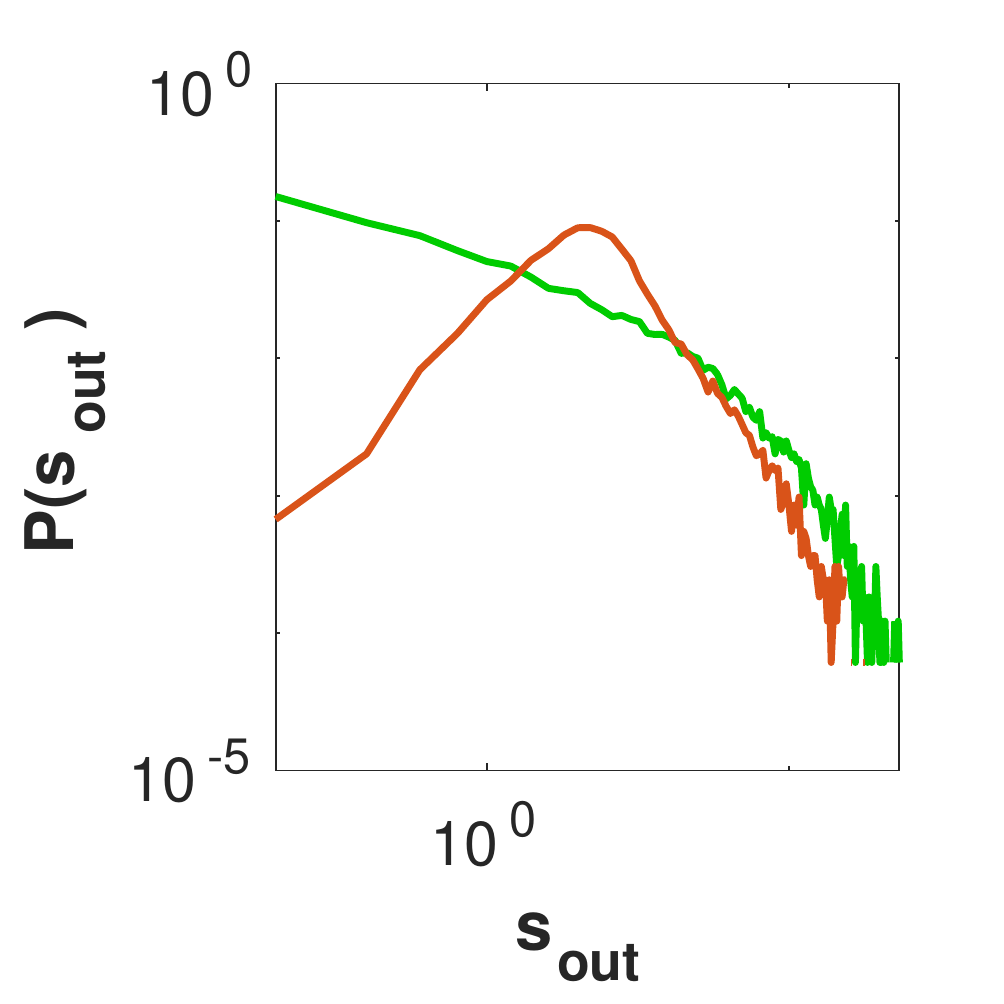} \includegraphics[width=0.23\linewidth]{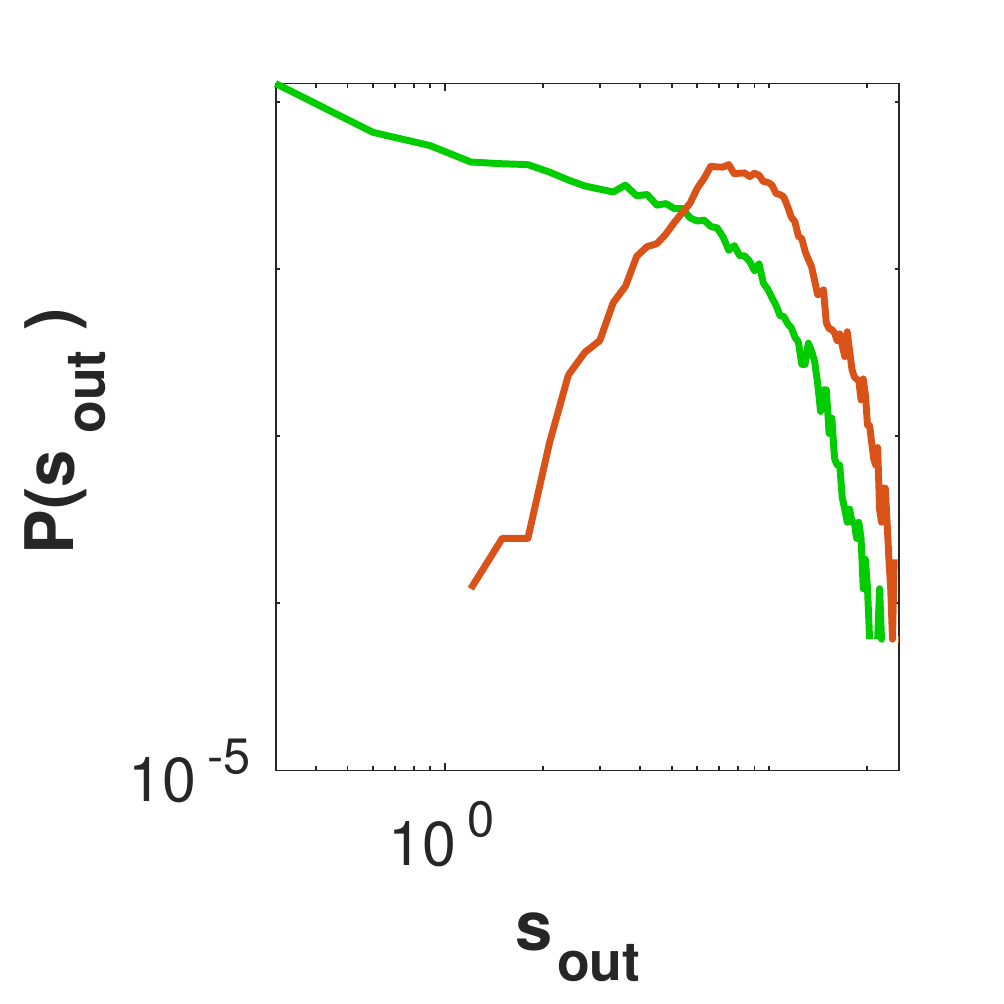}}
        
    \caption{\label{fig:distributions} Topological measures of a SSN $N^{r=4}$ obtained for vertices from each layer separately, according to the two input color-textures (RGB) (a).}
\end{figure}

The feature maps obtained from the characterization of SSN must be summarized in order to obtain a single and compact image descriptor. From a network science perspective, it is possible to obtain the probability distribution of the three centrality measures, which is a common approach for network characterization. We propose to compute the distribution for each layer of $N$, $W$ and $B$ separately, therefore, $z$ distributions are obtained from each network. It is intuitive to conclude that a separate analysis should provide more specific information regarding patterns occurring in each network layer. This technique improved our results if compared to using the whole network as proposed in \cite{scabini2019multilayer}. For the exact computation of the probability distribution function of each network layer, we fixed the number of bins for counting the input degree $k_{in}$ occurrence as the maximum possible degree. For the occurrence counting of the strength measure, the bin number is maximum possible degree multiplied by 10. We define the probability distribution function of each measure by $P^{k_{in}}$, $P^{s_{in}}$ and $P^{s_{out}}$. Figure \ref{fig:distributions} shows each distribution of a SSN $N^{r=4}$ according to the proposed layer-wise analysis, where it is possible to notice a clear distinction between the two different input textures. The topological arrangement of the network varies greatly according to the input texture and the layer being analyzed, however, it is possible to notice a power-law-like behavior in some cases. As a matter of comparison, the multilayer networks of \cite{scabini2019multilayer} seems to present a similar topology for different image inputs, with variations of the small-world effect and the occurrence of power-law-like degree distributions. On the other hand, here the directed SSN present heterogeneous arrangements where the network structure varies greatly between different texture, which then provides better topological measures for characterization. This difference between the previous work \cite{scabini2019multilayer} happens mostly due to the use of connection direction, the radially symmetric neighboring and the layer-wise network characterization.

\subsection{Feature vector}

Although the distributions $P^{k_{in}}$, $P^{s_{in}}$ and $P^{s_{out}}$ summarize the network topology, it is still impracticable to apply them as an image descriptor because as higher the parameter $r$, higher is the size of its combination, e.g. for a single network $N^{r=4}$ of a RGB image, the size would be (19 + 190 + 190)3 = 1197 (the size of each distribution obtained for each of the 3 image channels). Therefore, we propose to use statistical measures to further summarize the SSN structure. We employ the four measures proposed in \cite{scabini2019multilayer} (mean $\mu_f$, standard deviation $\sigma_f$, energy $e_f$, and entropy $\epsilon_F$) plus the third and fourth statistical moments, i.e. the Skewness
\begin{equation}
     \lambda_f = \frac{\frac{1}{|P^f|} \sum\limits_{i} (P^f(i) - \mu_f)^3}{\bigg(\sqrt{\frac{1}{|P^f|} \sum\limits_{i} (P^f(i) - \mu_f)^2}\bigg)^3}
\end{equation}
and the Kurtosis
\begin{equation}
    \kappa_f =  \frac{\frac{1}{|P^f|} \sum\limits_{i} (P^f(i) - \mu_F)^4}{\bigg(\sqrt{\frac{1}{|P^f|} \sum\limits_{i} (P^f(i) - \mu_f)^2}\bigg)^2}
\end{equation}

        

The combination of the six statistics for each of the three topological measures of each layer comprises a network descriptor. Consider a network $N^r=\{V_N, E_N\}$ and its set of vertices redefined as a combination of vertices from each of the $z$ layers $V_N = \{V_1,...,V_z\}$, then a vector $\varpi(V_i)_f$ represents the six statistics for a given layer $i$ and measure $f$. The concatenation of measures from each layer compose the network descriptor

 \begin{equation}\label{eq:fvlayers}
 \varpi(N^r) = [\varpi(V_1)_{k_{in}}, \varpi(V_1)_{s_{in}}, \varpi(V_1)_{s_{out}},..., \varpi(V_z)_{k_{in}}, \varpi(V_z)_{s_{in}}, \varpi(V_z)_{s_{out}}] 
 \end{equation}
 
 Considering the multiscale approach, we propose the use of a set of 1-by-1 increasing radius of $R=\{1,2,...,r_n\}$ to maintain the proportion of each symmetric neighborhood. Therefore, the final descriptor which addresses the dynamic evolution of each network is obtained by $ \varphi_N^{r_n} = [\varpi(N^{1}), ..., \varpi(N^{r_n})]$, $ \varphi_W^{r_n} = [\varpi(W^{1}), ..., \varpi(W^{r_n})]$ and $ \varphi_B^{r_n} = [\varpi(B^{1}), ..., \varpi(B^{r_n})]$. The combination of descriptors from networks $N$, $W$ and $B$ results in a complete representation that comprises the whole spatio-spectral information
 
 
 

  \begin{equation}\label{eq:fvm2}
 \varphi_{WB}^{r_n} = [\varphi_W^{r_n}, \varphi_B^{r_n}]
 \end{equation}
 
 \begin{equation}\label{eq:fvm2}
 \varphi^{r_n} = [\varphi_N^{r_n}, \varphi_W^{r_n}, \varphi_B^{r_n}]
 \end{equation}
 
\begin{figure}
    \centering
    \includegraphics[width=0.9\linewidth]{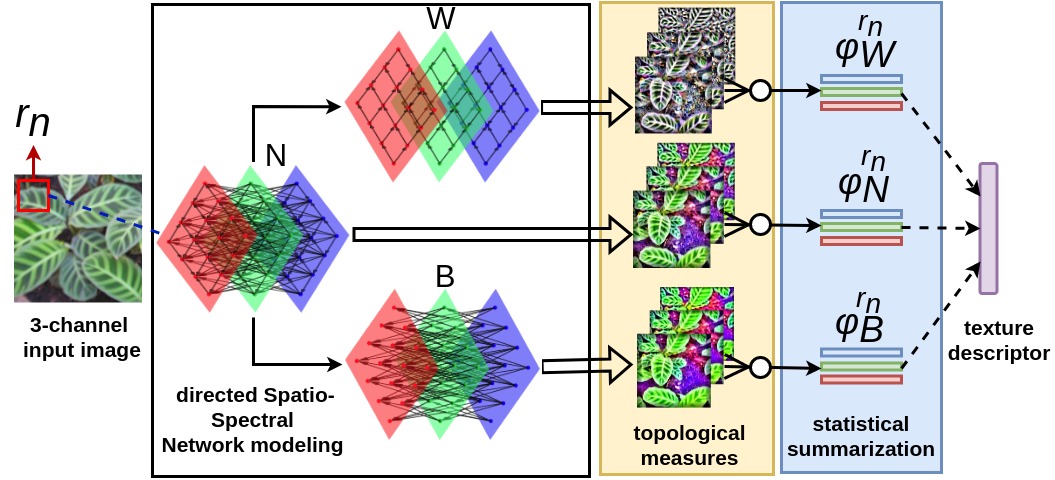}
    
    \caption{\label{fig:metodo2} The overall process of the proposed SSN approach given an input color image and the parameter $r_n$. The first step is the network modeling through the radially symmetric neighborhood, then the topological characterization is performed by vertex measures (input degree and input/output strength) from each layer separately, which are then summarized with statistical measures to compose the texture descriptor.}
\end{figure}

On Section \ref{sec:experiments} we present a deeper discussion regarding the feature vectors and its combinations. Figure \ref{fig:metodo2} illustrates each step of the whole process from the image input to the resulting feature vector obtained through the proposed SSN method.
 
\subsection{Computational complexity}

The cost of the proposed method is directly proportional to the image size and the largest radius used ($r_n$), which defines the window size of connection creation. The last radius, $r_n$, indicates that all pixels at distance $d \leq r_n$ must be visited to compose each neighborhood $G^{r_1}_{v_i}, ..., G^{r_n}_{v_i}$, according to the radially symmetric neighborhood previously defined. Therefore, consider an image $I$ with size $w*h$ and $z$ channels ($|I|=w*h*z$), and $g^{r_n} = (\frac{(2r+1)^2 -1}{2}z) +z$ as the number of pixels contained in all neighborhoods for a set of radius $R=\{1,...,r_n\}$. The cost of the image modelling as SSN is then

\begin{equation}\label{eq:cost}
O(|I|g^{r_n})
\end{equation}
considering that the modeling of $N$, $W$ and $B$ are performed together, as it is only necessary to verify the pixel channel in order to determine whether the connection belongs. As previously mentioned, the overall implementation of the proposed method is similar to the previous work \cite{scabini2019multilayer} (that has cost $O(2|I|g^{r_n})$) without the cost to compute the clustering coefficient of each vertex. Therefore, the network measures are computed during the modeling step with no additional cost, as there is no need to store the network. The characterization cost is then related to the computation of each measure distribution and its statistics, which is much smaller than the modeling cost. Therefore, the asymptotic limit of the proposed method is the same as defined in Equation \ref{eq:cost}. As the time-consuming experiment presented in \cite{scabini2019multilayer} shows, this approach is faster than recent deep convolutional networks.

\section{Experiments and Discussion}\label{sec:experiments}

This section presents the experiments performed to evaluate the proposed SSN under different scenarios and to compare our results with other methods from the literature. We perform a supervised classification scheme using the Linear Discriminant Analysis (LDA) classifier \cite{LDA}, which consists of finding a linear combination of characteristics where the variance between classes is greater than the intra-class variance. The performance is measured by the accuracy of leave-one-out cross-validation, which is a repetition of $n$ (number of samples of the dataset) train-test procedures where one sample is used for test and the remainder for training at each iteration (each sample is used as test once). The accuracy then is the percentage of correctly classified samples.

\subsection{Datasets}\label{sec:datasets}

The following color texture datasets from the literature are used:

\begin{itemize}
    
    \item \textbf{USPtex}: This dataset \cite{usptex} was built by the University of São Paulo and contains 191 classes of natural colored textures, found on a daily basis. The original images are 512x384 in size and are divided into 12 samples of size 128x128 without overlap, totaling 2292 images in total.

    \item \textbf{Outex13}: The Outex framework \cite{ojala2002outex} is proposed for the empirical evaluation of texture analysis methods. This framework consists of several different sets of images, and the Outex13 dataset (test suit Outex$\_$TC$\_$00013 on the web site \footnote{www.outex.oulu.fi}) focuses on the analysis of texture considering color as a discriminatory property. The dataset contains 1360 images divided into 68 classes, that is, 20 samples per class, of size 200x200.

    \item \textbf{MBT}: The Multi-Band Texture \cite{abdelmounaime2013newbrodatz} is composed of 154 colored images of classes formed by the combined effect of spatial variations within and between channels. This type of pattern appears in images with high spatial resolution, which are common in areas such as astronomy and remote sensing. Each of the 154 original images, 640x640 in size, is divided into 16 non-overlapping samples, size 160x160, composing a set of 2464 images.

    \item \textbf{CUReT}: The Columbia-Utrecht Reflectance and Texture dataset \cite{dana1999reflectance} is composed of colored images of materials. The base contains 61 classes with 92 samples each, where there is a wide variety of geometric and photometric properties as intra-class variations in rotation, illumination, and viewing angle. Classes represent surface textures of various materials such as aluminum foil, artificial grass, gypsum, concrete, leather, fabrics, among others.

\end{itemize}

\subsection{Proposed Method Analysis}\label{panalysis}

\begin{figure}[!htb]
    \centering
    \subfloat[USPtex]{\includegraphics[width=0.35 \linewidth]{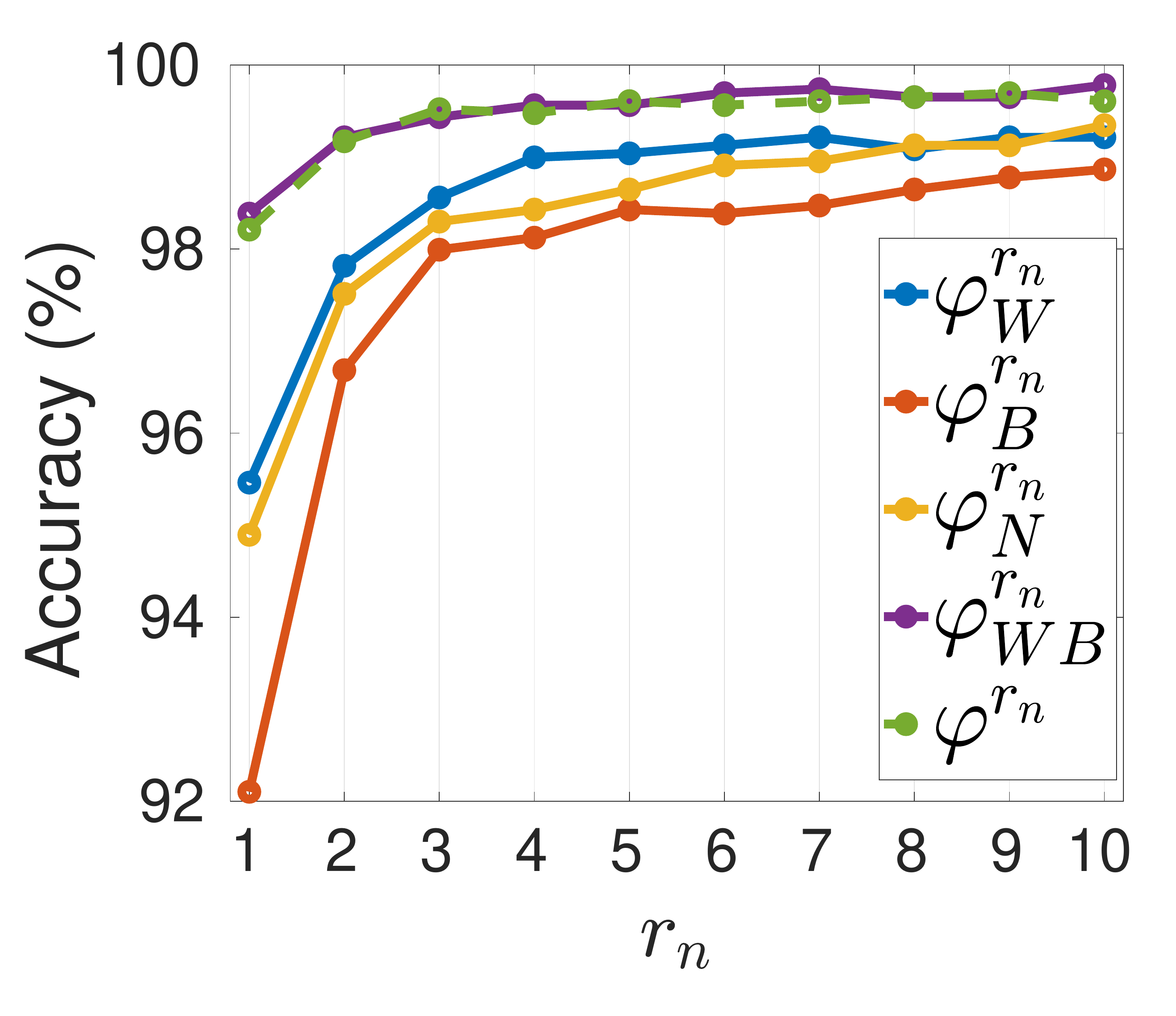}} \subfloat[Outex13]{\includegraphics[width=0.35 \linewidth]{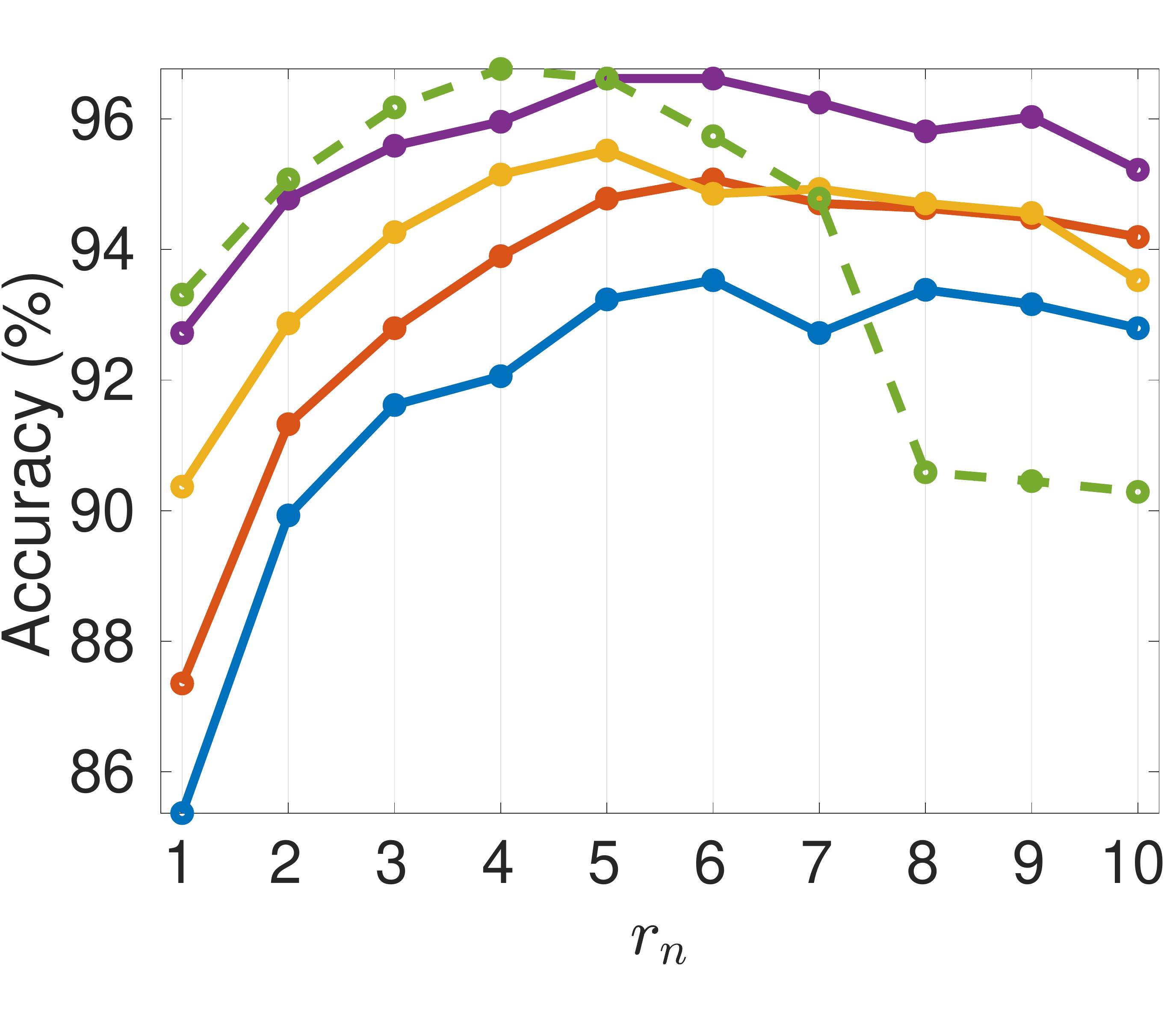}} \\
    \subfloat[MBT]{\includegraphics[width=0.35 \linewidth]{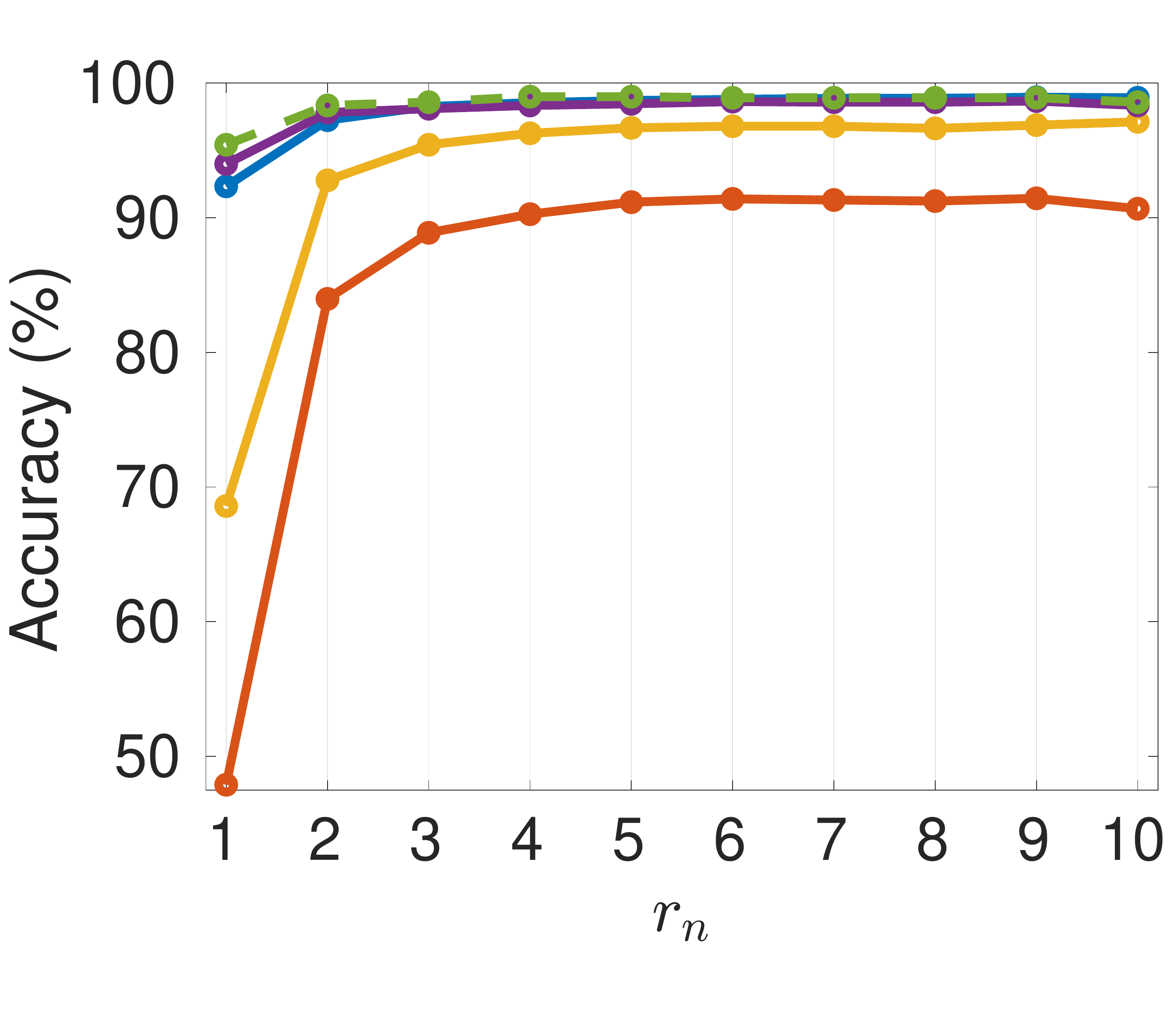}} \subfloat[CUReT]{\includegraphics[width=0.35 \linewidth]{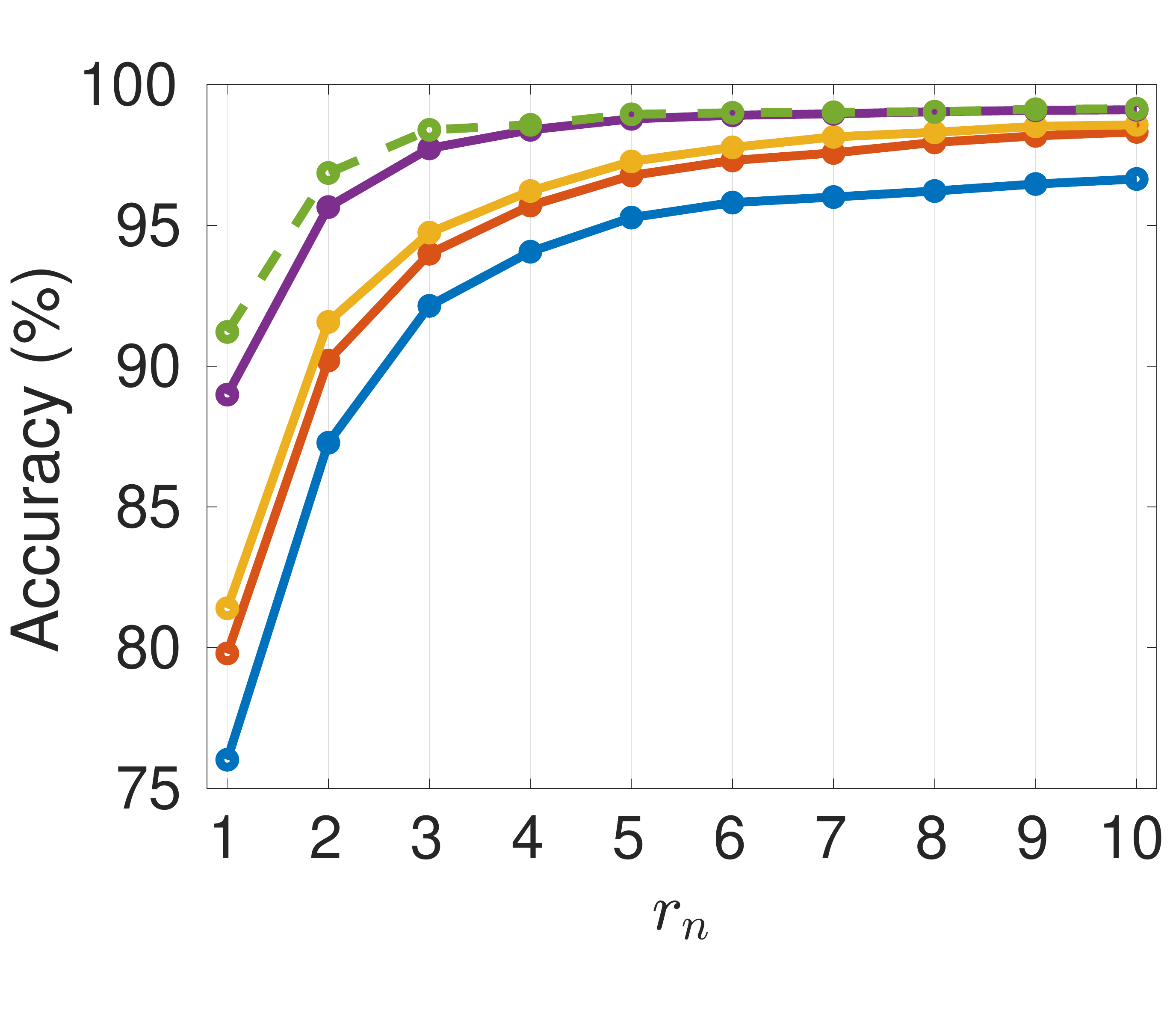}}
        
    \caption{\label{fig:pmanalysis} Performance of descriptors obtained from each network $N$, $W$ and $B$ (separated or combined), by varying the parameter $r^n$ on each dataset.}
\end{figure}

We analyze the proposed method in terms of its parameter $r_n$ and the impact on using different feature vectors from the combination of the networks $W^{r_n}$, $B^{r_n}$ and $N^{r_n}$. Figure \ref{fig:pmanalysis} shows how the accuracy rate obtained with each feature vector $\varphi_{W}^{r_n}$, $\varphi_{B}^{r_n}$, $\varphi_{N}^{r_n}$, $\varphi_{WB}^{r_n}$ and $\varphi^{r_n}$ changes as $r_n$ increases. It is possible to notice that the performance grows to a certain point and then stabilizes, except for Outex13 where it starts to drop after $r_n=6$. This indicates that smaller values of $r_n$ are not sufficient to achieve a complete multiscale texture characterization, and also that too higher values on Outex13 can provide feature redundancy as the number of descriptors grows. On the other hand, peaks of performances are achieved with $r_n=10$ on USPtex and CUReT, but with a small performance difference in comparison to smaller $r_n$ values. Therefore, a more stable region lies around $3 \leq r_n \leq 6$ if we consider all datasets. Regarding the feature vectors, it is possible to notice that the performance for each network alone vary between datasets, where $\varphi_{W}^{r_n}$ is better on USPtex and MBT, while $\varphi_{N}^{r_n}$ is better on Outex13 and CUReT. On the other hand, both vectors $\varphi_{WB}^{r_n}$ and $\varphi^{r_n}$ present the highest results in most cases, which is to be expected as it combines measures from both within and between-channel analysis. We then suggest the use of the combinatorial feature vectors in order to obtain better results in different scenarios.

On Table \ref{table:averagepf} we show the average accuracy (for the 4 datasets) of the combinatorial feature vectors $\varphi_{WB}^{r_n}$ and $\varphi^{r_n}$ as $r_n$ increases, along with the corresponding number of descriptors. As we previously suggested, the interval $3 \leq r_n \leq 6$ has a smaller standard deviation, which indicates a more stable region. Moreover, the highest average performance is obtained with $r_n=6$ for $\varphi_{WB}^{r_n}$ using 648 descriptors, and with $r_n=4$ and $5$ for $\varphi^{r_n}$ using 648 or 810 descriptors, respectively. On this context, the results suggest that the 2 alternatives $\varphi_{WB}^{6}$ or $\varphi^{4}$ has a similar average performance than $\varphi^{5}$ but using a smaller number of descriptors. For a more satisfactory performance in most scenarios, we suggest the use of $\varphi_{WB}^{6}$ due to the use of a larger neighborhood, providing a better multiscale analysis.

\begin{table}[!htb]                                                                                                                
    \centering                                                                                                                       
    \caption{Mean accuracy over the four datasets (standard deviation in brackets) using different feature vectors.}                                                                                                      
    \label{table:averagepf}                 
    \begin{tabular}{|c|cc|cc|}  
    \hline
    & \multicolumn{2}{|c|}{$\varphi_{WB}^{r_n}$} & \multicolumn{2}{|c|}{$\varphi^{r_n}$} \\ 
    $r_n$ & descriptors & mean acc. & descriptors & mean acc. \\
    \hline
      1&   108   &93.5($\pm3.4$)& 162    &94.5($\pm2.6$)\\
      \hline
      2&   216   &96.9($\pm1.7$)& 324     &97.4($\pm1.6$)\\
      \hline
      3&    324  &97.7($\pm1.4$)&486    &98.2($\pm1.2$)\\
      \hline
      \textbf{4}&   432   &98.1($\pm1.3$)&648    &\textbf{98.5}($\pm1.0$)\\
      \hline
      \textbf{5}&   540   &98.4($\pm1.1$)&810    &\textbf{98.5}($\pm1.1$)\\
      \hline
      \textbf{6}&   648   &\textbf{98.5}($\pm1.1$)&    972&98.3($\pm1.5$)\\
      \hline
      7&   756   &98.4($\pm1.3$)&1134    &98.1($\pm1.9$)\\
      \hline
      8&   864   &98.3($\pm1.5$)&1296    &97.0($\pm3.7$)\\
      \hline
      9&    972  &98.4($\pm1.4$)&1458    &97.4($\pm3.2$)\\
      \hline
     10&   1080   &98.1($\pm1.7$)&1620    &97.4($\pm3.0$)\\
     \hline
    \end{tabular}                           
\end{table}

\subsection{Comparison with the literature}

The following literature methods are considered for comparison: Opponent-Gabor \cite{opponentgabor1998multiscale} (264 descriptors), Local Phase Quantization (LPQ) \cite{ojansivu2008blur} (integrative, 768 descriptors), Complete Local Binary Patterns (CLBP) \cite{guo2010completed} (integrative, 354 descriptors), Complex Networks Traditional descriptors (CNTD) \cite{backes} (108 descriptors for grayscale and 324 for integrative) and Multilayer Complex Network Descriptors (MCND) \cite{scabini2019multilayer} (we use the best results available in the paper). 
We also compare results with some well-known deep convolutional neural networks (DCNN), these are: AlexNet \cite{krizhevsky2012imagenet}, VGG16 and VGG19 \cite{simonyan2014very}, InceptionV3 \cite{szegedy2015going,szegedy2016rethinking}, ResNet50 and ResNet101 \cite{he2016deep}. The DCNN are employed as feature extractors using a pre-trained architecture on the ImageNet object recognition dataset, from the Large Scale Visual Recognition Challenge \cite{imagenet}. Each considered architecture and its pre-trained weights are available at the Neural Network Toolbox of the Matlab 2018a$^{tm}$ software \footnote{\url{https://www.mathworks.com/help/deeplearning/ug/pretrained-convolutional-neural-networks.html}}. The use of pre-trained DCNN achieves higher performance than training these models from scratch on the considered color texture datasets, as results in \cite{scabini2019multilayer} shows. As each DCNN has a fixed input size, images from each dataset are resized before being processed, and the classification part (fully-connected layers) is removed. Therefore the 2D feature activation maps generated by the last convolutional layers are considered to compute image features, this approach has been used previously \cite{lin2013network,cimpoi2014describing,cimpoi2016}. To obtain a feature vector a Global Average Pooling (GAP) is applied as in \cite{lin2013network}, producing a vector of a size corresponding to the number of 2D feature activation maps. However, most of the last convolutional layer of DCNNs produce a high number of feature activation maps such as 2048 for the Inception and ResNet models. In this sense, we propose to use previous convolutional layers until the size is smaller than 800. This reduction is performed for various reasons, where the most obvious is to avoid the curse of dimensionality, caused by the exponential increase in the volume associated to the addition of extra dimensions to the Euclidean space \cite{keogh2017curse}. Another reason is a known effect of the LDA classifier, the Singularity, Small Sample Size (SSS), which happens when there is high dimensional feature vectors or a small number of training images \cite{huang2002solving,tharwat2017linear}. It is also important to highlight that we use the raw output of the corresponding convolutional layers, i.e we do not apply its following summarizing functions such as ReLU and local poolings. This approach improves the average performance of DCNN as texture feature extractors when compared to the results in \cite{scabini2019multilayer}, therefore we believe that it is reasonable for our purpose. Additionally to the aforementioned situations, the reduction of the DCNN feature vector also improves the computational cost of the training and classification step, while keeping a reasonable number of descriptors in comparison to other hand-crafted descriptors. The exact number of features obtained with InceptionV3 and both ResNet models are, respectively, 768 and 512. For AlexNet and both VGG models, the size corresponds to its last convolutional layer, which is 256 and 512, respectively.

We compare the accuracy rate between the proposed approach and the aforementioned methods on the four considered datasets, results are shown on Table \ref{tab:comparisons}, the highest result of each column is highlighted in bold type. For the USPtex dataset, the lowest results are achieved by the integrative methods LPQ, CLBP and CNTD (around $90\%$ and $98\%$), followed by the MCND and the Opponent-Gabor descriptors (around $99\%$). DCNNs performs near the highest results, with small differences between each model. The highest result is obtained using the proposed method and $\varphi_{WB}^{10}$, achieving $99.8\%$ of accuracy rate. The suggested feature vector $\varphi_{WB}^{6}$ achieves the second highest result of $99.7\%$, equivalent to the performance obtained by the ResNet50 DCNN. On the other hand, DCNN performs better than the previous hand-crafted descriptors, but are overcome by the proposed SSN.

\begin{table}[!htb]
    \centering
    \caption{\label{tab:comparisons} Accuracy rate of leave-one-out cross-validation with the LDA classifier using the proposed approach and several literature methods on four color texture datasets. The last column represents the average and standard deviation of each method over all the datasets.}
    \begin{tabular}{c|c|c|c|c|c}
        \hline
         Method &  USPtex & Outex13 & MBT & CUReT & Average\\
        \hline

     Opponent-Gabor (1998)&99.1&93.5&97.6&95.8& 96.5($\pm2.1$)\\
     LPQ integrative (2008)&90.4&80.1&95.7&91.7& 89.5($\pm5.8$) \\
     CLBP integrative (2010)&97.4&89.6&98.2&91.8& 94.3($\pm3.6$)\\

     CNTD (grayscale) (2013) & 92.3 &  86.8 & 83.7&84.2 & 86.8($\pm3.4$)\\
     CNTD integrative (2013)&97.9&92.3&98.5&91.9& 95.2($\pm3.1$)\\
     MCND (2019) \cite{scabini2019multilayer}&99.0 &95.4 &97.0 & 97.1& 97.1($\pm1.3$)\\
     
     \hline
      AlexNet (2012)&99.6&91.4&97.8&98.2& 96.8($\pm3.2$)\\
     VGG16 (2014) &99.5&91.1&97.2&98.5& 96.6($\pm3.3$)\\
     VGG19 (2014) &99.5&90.7&96.3&98.6& 96.3($\pm3.4$)\\
     InceptionV3 (2016) &99.5&89.5&94.4&97.3&  95.2($\pm3.7$)\\
     ResNet50 (2016) &99.7&91.5&94.9&98.7& 96.2($\pm3.3$) \\
     ResNet101 (2016)&99.5&91.3&94.6&98.8& 96.1($\pm3.3$)\\
     
    \hline
    \textbf{SSN ($\varphi_{WB}^{6}$)}& 99.7&96.6&98.6&98.9& \textbf{98.5}($\pm1.1$)\\
    \textbf{SSN ($\varphi_{WB}^{10}$)}& \textbf{99.8}&95.2&98.3&\textbf{99.1}& 98.1($\pm1.7$)\\
    
      \textbf{SSN ($\varphi^{4}$)}& 99.5&\textbf{96.8}&\textbf{99.0}&98.6& \textbf{98.5}($\pm1.0$)\\
     \textbf{SSN ($\varphi^{5}$)}&99.6&96.6&\textbf{99.0}&98.9& \textbf{98.5}($\pm1.1$)\\

     \hline
        \end{tabular}
    \end{table}

Outex13 is the hardest dataset in terms of color texture characterization, as results shows. Moreover, on this dataset the performance of DCNNs drops considerably, performing around $90.7\%$ and $91.5\%$. The integrative LPQ and CLBP and the grayscale approach of CNTD present the lowest accuracies, with $80.1\%$, $89.6\%$ and $86.8\%$, respectively. On the other hand, the integrative CNTD method performs above the DCNN ($92.3\%$), and the Opponent-Gabor method overcomes both the integrative and the DCNN methods, achieving $93.5\%$. The highest performance is obtained by the proposed SSN and the MCND, with $96.8\%$ and $95.4\%$, respectively. The suggested feature vector $\varphi_{WB}^{6}$ achieves $96.6\%$, which also overcomes the other methods, where the performance improvement is of $1.2\%$ over MCND and $5.1\%$ over the best DCNN, ResNet50. The results of the best methods on this dataset, Opponent-Gabor, MCND, and SSN, corroborates to the importance of CN and the within-between channel analysis for color texture characterization.

On the MBT dataset, we see again a different performance pattern regarding the literature approaches. Firstly, the large performance difference between the grayscale and the integrative CNTD method ($83.7\%$ and $98.5\%$) indicates that color plays an important role in the MBT color texture characterization. The DCNN AlexNet and VGG16 achieve around $97\%$ of accuracy, similar to the Opponent-Gabor approach. On the other hand, the deeper DCNN (Inception and ResNet) achieves the lowest results, around $94\%$. The highest accuracy is obtained by the proposed method, with $99\%$ at its best configuration and $98.6\%$ using the suggested feature vector $\varphi_{WB}^{6}$.

The last dataset, CUReT, is the largest in terms of the number of samples per classes, and we can also notice a different performance pattern regarding the literature methods. On this dataset, the depth of DCNNs seems to be beneficial to performance, as we can observe for VGG and ResNet, which can be related to the dataset size. The integrative methods present the lowest performance, around $92\%$, performing below Opponent-Gabor and MCND ($95.8\%$ and $97.1\%$, respectively). The DCNN descriptors perform above the previous methods, where the deeper network analyzed, ResNet101, achieves $98.8\%$. The proposed method again overcomes the other approaches, where the highest accuracy rate is achieved with the configuration $\varphi_{WB}^{10}$ ($99.1\%$), and the proposed configuration $\varphi_{WB}^{6}$ achieves $98.9\%$.

The last column of Table \ref{tab:comparisons} shows the average accuracy of each method over all datasets (standard deviation in brackets). We can see that the proposed SSN achieves the highest average performance for any configuration, and also the lower standard deviations, corroborating to its robustness for different color texture scenarios. The DCNN overcomes the previous hand-crafted descriptors, however, its high standard deviation reflects its performance drop on the Outex13 and MBT datasets. Similarly, the integrative descriptors also present a varying performance, as its standard deviation shows. Regarding the average performance of the Opponent-Gabor method, we can conclude that the opponent color processing allows a better characterization in some cases, therefore this method has a higher performance than integrative methods and performs close to DCNNs. This corroborates to the benefits of the opponent color analysis from which the proposed method benefits, however, our approach keeps a higher performance and a lower oscillation between datasets in comparison to the Opponent-Gabor method. Considering the CN-based methods, we can see that the CNTD approach also has an oscillating performance, and is overcome by the MCND which has the second highest average performance and also the second lowest standard deviation, behind only the proposed method. Overall, the results presented here corroborate to the effectiveness of SSN for color texture analysis in various scenarios, where it keeps the highest performance in all cases, while most of the other literature methods oscillate.

\subsection{Performance under different color spaces}

In color texture analysis the image representation through different color spaces may impact the characterization performance. On this context, this section presents a set of experiments by varying the image color space of the four considered datasets and comparing the performance between the proposed method ($\varphi_{WB}^{6}$) and other literature approaches. The MCND approach is not included in this analysis because the available results concern only the best configuration of the method for each dataset separately in the RGB space. Four color spaces are considered for analysis, being one for each different approach: RGB, LAB, HSV, and $I_1I_2I_3$. These techniques are described in depth at Section \ref{sec:color}. As some color spaces use different value ranges for different channels, we normalize its values to $[0,1]$. First, we analyze the performance for the USPtex and Outex13 datasets, the accuracies of leave-one-out cross-validation with the LDA classifier are shown in Table \ref{tab:cor1}. It is possible to observe how different color spaces influence each method, in different ways. For example, the integrative methods, had increased performance in the HSV space, while the opposite happens with the DCNN. In fact, this effect on the convolutional networks also happens in all other channels, albeit with a lower intensity. This is somewhat expected since these networks are pre-trained in a set of RGB images. The two methods that presented greater robustness to the different types of color spaces were, respectively, the proposed SSN and the Opponent-Gabor method. This indicates that the opponent color processing approach allows a greater tolerance to the different ways of representing the colors of the images. The proposed method incorporates these characteristics and, at the same time, increases the invariance to color spaces, reaching the highest average performance and the lowest standard deviation of $99.6(\pm0.1)$ versus $98.8(\pm0.6)$ of Opponent-Gabor. In addition, the SSN method also achieves the highest results in each color space individually.

\begin{table}[!htb]
    \centering
    \caption{\label{tab:cor1} Performance on the USPtex and Outex datasets with different color spaces, the last column represents the average performance and standard deviation over all color spaces on the corresponding dataset.}
    
    \begin{tabular}{cc|c|c|c|c|c}
        \hline
        &Method & RGB & LAB & HSV & $I_1I_2I_3$ & Average\\
        \hline
        \multirow{11}{*}{\textbf{USPtex}}&LPQ i.&90.4&95.0&96.6&94.6&94.1($\pm2.7$)\\
        &CLBP i.&97.4&98.6&98.6&98.5&98.3($\pm0.6$)\\
        &Opponent-Gabor&99.1&99.0&97.9&99.3&98.8($\pm0.6$)\\
        &CNTD i.&97.9&98.3&99.1&98.2&98.4($\pm0.5$)\\
        
        &AlexNet&99.6&99.0&94.7&99.3&98.2($\pm2.3$)\\
        &VGG16&99.5&98.6&94.4&98.7&97.8($\pm2.3$)\\
        &VGG19&99.5&98.2&92.1&98.6&97.1($\pm3.4$)\\
        &InceptionV3&99.5&97.7&94.7&98.1&97.5($\pm2.0$)\\
        &ResNet50&\textbf{99.7}&98.2&92.5&98.5&97.2($\pm3.2$)\\
        &ResNet101&99.5&98.0&92.8&98.5&97.2($\pm3.0$)\\

        
        &\textbf{SSN ($\varphi_{WB}^{6}$)}&\textbf{99.7}&\textbf{99.7}&\textbf{99.4}&\textbf{99.4}&\textbf{99.6}($\pm0.1$)\\

    \hline

        \multirow{11}{*}{\textbf{Outex13}}&LPQ i.&80.1&74.8&78.2&76.0&77.3($\pm2.4$)\\
        &CLBP i.&89.6&86.8&88.2&86.6&87.8($\pm1.4$)\\
        &Opponent-Gabor&93.5&91.3&91.7&91.3&91.9($\pm1.1$)\\
        &CNTD i.&92.3&90.6&94.0&90.6&91.9($\pm1.7$)\\
        
        &AlexNet&91.4&91.6&91.0&91.1&91.3($\pm0.3$)\\
        &VGG16&91.1&89.2&87.6&89.8&89.4($\pm1.4$)\\
        &VGG19&90.7&89.1&85.9&90.8&89.1($\pm2.3$)\\
        &InceptionV3&89.5&85.0&84.2&85.3&86.0($\pm2.4$)\\
        &ResNet50&91.5&88.7&86.8&89.0&89.0($\pm1.9$)\\
        &ResNet101&91.3&90.1&86.4&88.8&89.2($\pm2.1$)\\
        
        
        &\textbf{SSN ($\varphi_{WB}^{6}$)}&\textbf{96.6}&\textbf{94.3}&\textbf{95.7}&\textbf{94.6}&\textbf{95.3}($\pm1.1$)\\
        
        
    \hline
    
\end{tabular}
\end{table}

In the Outex13 dataset, the methods with the lowest performance are, respectively, LPQ and InceptionV3. The other DCNN, except for AlexNet, get an average of around $89\% $. The CLBP method has its performance reduced in the LAB and $ I_1I_2I_3$ color spaces, which results in an average below the convolutional networks, of $87.8(\pm 1.4)$. The integrative CNTD method overcomes other integrative methods, obtaining an average of $91.9\%$, a result also obtained approximately by the Opponent-Gabor method and by the AlexNet DCNN. This shows that traditional integrative methods have limitations on this dataset when different color spaces are considered. For example, the CLBP method, in the USPtex dataset, is benefited by the changes of color space, which does not happen in Outex13. The methods with the greatest robustness to the different color spaces in this dataset are again the proposed SSN and the Opponent-Gabor method, with an average performance of $95.3(\pm1.1)$ and $91.9(\pm1.1)$, respectively. In this context and considering the results obtained by the Opponent-Gabor and the integrative CNTD methods, it is possible to note that CN and opponents techniques present the best performances on this dataset, and the SSN method which combines these properties achieves the highest results for all color spaces.

Table \ref{tab:cor2} shows the obtained results for the MBT and CUReT datasets. Regarding MBT, the highest results are $97.3(\pm1.3)$ of the SSN method, $96.6(\pm2.3)$ of the integrative CNTD approach and $96.6$ obtained by both Opponent-Gabor and CLBP methods. On this dataset, the average performance of the deepest neural networks (InceptionV3 and both ResNets) is the lowest observed, close to $92\% $, while the smaller networks AlexNet and VGG16 obtain, respectively, $ 95.6 (\pm4.8) $ and $ 94.7(\pm4.2) $. The high standard deviation of the DCNN is due to the sharp performance drop with the HSV color space, with losses of almost $10\%$. Differently, from what is observed in the other bases, this also happens with all other methods, but with a smaller intensity if compared to the DCNN. On this aspect the proposed method proved to be more robust, achieving the highest performance under HSV and also the lowest variation between color spaces. The results obtained in the CUReT dataset corroborate the robustness of the proposed method to different color spaces. The highest result is obtained by SSN in all color spaces with an average performance of $98.7(\pm0.2)$. On this dataset, the DCNN ResNet (50 and 101) outperform the integrative methods in the individual LAB and $ I_1I_2I_3 $ color spaces, and despite presenting again a loss in the HSV space, their average performance overcomes the other literature methods. In the HSV space, the best results are obtained by the proposed method and by the Opponent-Gabor method. In general, on this dataset both integrative and convolutional networks have the highest standard deviation, indicating that their performance oscillates more between different color spaces. The methods with lower oscillation of performance are the proposed SSN and the Opponent-Gabor method, respectively, reinforcing the robustness of the opponent color processing and the CN-based approach.

\begin{table}[!htb]
    \centering
    \caption{\label{tab:cor2} Performance on the MBT and CUReT datasets with different color spaces, the last column represents the average performance and standard deviation over all color spaces on the corresponding dataset.}
    
    \begin{tabular}{cc|c|c|c|c|c}
    \hline
        &Method & RGB & LAB & HSV & $I_1I_2I_3$ & Average\\
        \hline

        \multirow{11}{*}{\textbf{MBT}}&LPQ i.&95.7&94.2&90.6&93.9&93.6($\pm2.1$)\\
        &CLBP i.&98.2&95.9&94.6&96.6&96.3($\pm1.5$)\\
        &Opponent-Gabor&97.6&96.9&93.9&96.7&96.3($\pm1.6$)\\
        &CNTD i.&98.5&96.7&93.3&97.8&96.6($\pm2.3$)\\
        
        &AlexNet&97.8&\textbf{98.1}&88.4&\textbf{98.2}&95.6($\pm4.8$)\\
        &VGG16&97.2&97.0&88.5&96.3&94.7($\pm4.2$)\\
        &VGG19&96.3&95.8&87.9&95.8&94.0($\pm4.0$)\\
        &InceptionV3&94.4&95.0&85.8&93.1&92.1($\pm4.3$)\\
        &ResNet50&94.9&94.2&85.1&93.6&91.9($\pm4.6$)\\
        &ResNet101&94.6&94.1&85.3&92.7&91.6($\pm4.3$)\\
        
        &\textbf{SSN ($\varphi_{WB}^{6}$)}&\textbf{98.6}&97.1&\textbf{95.6}&98.1&\textbf{97.3}($\pm1.3$)\\

        \hline 
        \multirow{11}{*}{\textbf{CUReT}}&LPQ i.&91.7&94.8&96.3&93.0&93.9($\pm2.0$)\\
        &CLBP i.&91.8&94.2&96.2&93.9&94.0($\pm1.8$)\\
        &Opponent-Gabor&95.8&95.6&97.4&95.3&96.0($\pm0.9$)\\
        &CNTD i.&91.9&92.7&95.4&91.2&92.8($\pm1.8$)\\
        
        &AlexNet&98.2&96.6&92.9&96.8&96.1($\pm2.2$)\\
        &VGG16&98.5&97.6&93.7&97.4&96.8($\pm2.1$)\\
        &VGG19&98.6&97.9&92.3&97.5&96.6($\pm2.9$)\\
        &InceptionV3&97.3&95.5&89.4&95.0&94.3($\pm3.4$)\\
        &ResNet50&98.7&97.3&93.9&97.8&96.9($\pm2.1$)\\
        &ResNet101&98.8&98.0&93.4&97.9&97.0($\pm2.5$)\\
        
        
        &\textbf{SSN ($\varphi_{WB}^{6}$)}&\textbf{98.9}&\textbf{98.4}&\textbf{98.9}&\textbf{98.6}&\textbf{98.7}($\pm0.2$)\\
        \hline
        
    \end{tabular}
\end{table}

\begin{figure}
    \centering
    \includegraphics[width=0.7\linewidth]{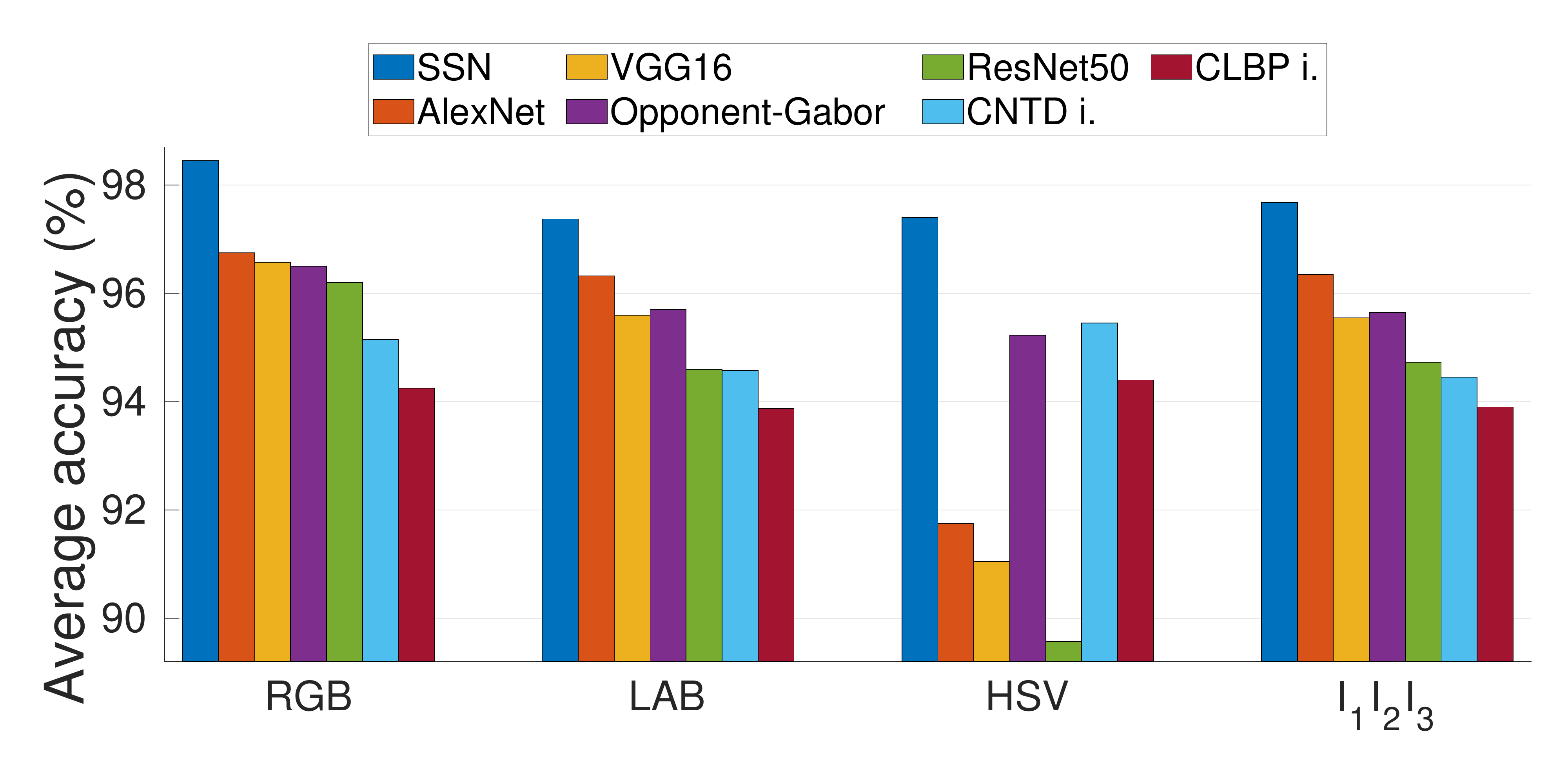}
    \caption{ \label{fig:robustness} Average performance of the 7 best methods over all the 4 datasets for each color space separately.}
\end{figure}

To analyze the overall performance of the methods on each color space separately, we compute the average accuracy over all the datasets, results are shown in Figure \ref{fig:robustness} for the 7 best methods. It is possible to notice that SSN has the highest average performance for all color spaces. The DCNN AlexNet achieves the second best performance overcoming other compared methods for the RGB, LAB and $I_1I_2I_3$ color spaces, but performs poorly with the HSV color space, along with the other DCNN. On the other hand, the proposed method overcomes the other approaches with the largest margin on the HSV color space.

The color space experiment shows that the proposed SSN has great robustness both between datasets or between different color spaces for a single dataset. In all scenarios, its performance variation is significantly smaller in comparison to the other literature methods, while also achieving the highest results. Therefore, considering all the obtained results, SSN stands out as a very effective approach for color texture characterization in a wide range of different scenarios, such as under different color texture properties (considering the heterogeneity of the datasets) and different color space representations.

\section{Conclusion}

This work introduces a new single-parameter technique for color texture analysis which consists of the modeling and characterization of a directed Spatio-Spectral Network (SSN) from the image color channels. Each pixel, in each channel, is considered as a network vertex, resulting in a multilayer structure. The connections are created according to a proposed technique of radially symmetric neighborhood given a distance limiting parameter $r_n$. Directed connections within-between color channel are defined pointing towards the pixels of higher intensity, and the connection weight consists of a normalized calculation of the intensity difference multiplied by the pixel Euclidean distance. This process results in a rich network with deep spatio-spectral properties, capturing a wide range of texture patterns related to the combined spatial and spectral intensity variations. This network is then quantified with topological measures following a complete characterization procedure, considering different measures, different connection types and analyzing each network layer separately. The whole process results in a compact and effective image descriptor for color texture characterization.

We performed classification experiments on four datasets (USPtex, Outex13, MBT, and CUReT) in order to evaluate the proposed SSN and to compare its performance to other methods from the literature. 12 methods are considered for comparison, including Gabor filters obtained from opponent color channels, integrative versions of grayscale texture descriptors (LPQ, CLBP, and CNTD), CN-based methods (CNTD and MCND) and DCNN (AlexNet, VGG, Inception, and ResNet). Results show that the proposed approach has higher performance for all datasets, and also the smallest variation between datasets, corroborating to its robustness to different scenarios. We also perform an evaluation of the impacts of different color spaces (RGB, LAB, HSV and $I_1I_2I_3$) on color texture characterization, which shows that SSN also has the higher average performance and higher tolerance to each color space than the other compared methods from the literature. The obtained results suggest that the spatio-spectral approach combined with the flexibility of complex networks to model and characterize real phenomena is a powerful technique for color texture analysis, and its properties should be further explored.

\section*{Acknowledgments}

L. F. S. Scabini acknowledges support from CNPq (Grants \#134558/2016-2 and \#142438/2018-9). L. C. Ribas gratefully acknowledges the financial support grant \#2016/23763-8 and \#2016/18809-9, S\~ao Paulo Research Foundation (FAPESP). O. M. Bruno acknowledges support from CNPq (Grant \#307897/2018-4) and FAPESP (grant \#2014/08026-1 and 2016/18809-9). The authors are also grateful to the NVIDIA GPU Grant Program for the donation of the Quadro P6000 and the Titan Xp GPUs used on this research.


\begin{thebibliography}{10}

\bibitem{abdelmounaime2013newbrodatz}
S.~Abdelmounaime and H.~Dong-Chen.
\newblock New brodatz-based image databases for grayscale color and multiband
  texture analysis.
\newblock {\em ISRN Machine Vision}, 2013, 2013.

\bibitem{fourier}
R.~Azencott, J.-P. Wang, and L.~Younes.
\newblock Texture classification using windowed fourier filters.
\newblock {\em Pattern Analysis and Machine Intelligence, IEEE Transactions
  on}, 19(2):148--153, 1997.

\bibitem{usptex}
A.~R. Backes, D.~Casanova, and O.~M. Bruno.
\newblock Color texture analysis based on fractal descriptors.
\newblock {\em Pattern Recognition}, 45(5):1984--1992, 2012.

\bibitem{backes}
A.~R. Backes, D.~Casanova, and O.~M. Bruno.
\newblock Texture analysis and classification: A complex network-based
  approach.
\newblock {\em Information Sciences}, 219:168--180, 2013.

\bibitem{scalefreeCN}
A.-L. Barab{\'a}si and R.~Albert.
\newblock Emergence of scaling in random networks.
\newblock {\em science}, 286(5439):509--512, 1999.

\bibitem{bu2019deep}
X.~Bu, Y.~Wu, Z.~Gao, and Y.~Jia.
\newblock Deep convolutional network with locality and sparsity constraints for
  texture classification.
\newblock {\em Pattern Recognition}, 91:34--46, 2019.

\bibitem{cantero2018importance}
S.~V. A.~B. Cantero, D.~N. Gon{\c{c}}alves, L.~F. dos Santos~Scabini, and W.~N.
  Gon{\c{c}}alves.
\newblock Importance of vertices in complex networks applied to texture
  analysis.
\newblock {\em IEEE transactions on cybernetics}, 2018.

\bibitem{casanova2016}
D.~Casanova, J.~Florindo, M.~Falvo, and O.~Bruno.
\newblock Texture analysis using fractal descriptors estimated by the mutual
  interference of color channels.
\newblock {\em Information Sciences}, 346:58--72, 2016.

\bibitem{cernadas2017}
E.~Cernadas, M.~Fern{\'a}ndez-Delgado, E.~Gonz{\'a}lez-Rufino, and
  P.~Carri{\'o}n.
\newblock Influence of normalization and color space to color texture
  classification.
\newblock {\em Pattern Recognition}, 61:120--138, 2017.

\bibitem{chalumeau2006}
T.~Chalumeau, L.~d.~F. Costa, O.~Laligant, and F.~Meriaudeau.
\newblock Optimized texture classification by using hierarchical complex
  network measurements.
\newblock In {\em Electronic Imaging 2006}, pages 60700Q--60700Q. International
  Society for Optics and Photonics, 2006.

\bibitem{cimpoi2014describing}
M.~Cimpoi, S.~Maji, I.~Kokkinos, S.~Mohamed, and A.~Vedaldi.
\newblock Describing textures in the wild.
\newblock In {\em Proceedings of the IEEE Conference on Computer Vision and
  Pattern Recognition}, pages 3606--3613, 2014.

\bibitem{cimpoi2016}
M.~Cimpoi, S.~Maji, I.~Kokkinos, and A.~Vedaldi.
\newblock Deep filter banks for texture recognition, description, and
  segmentation.
\newblock {\em International Journal of Computer Vision}, 118(1):65--94, 2016.

\bibitem{aplicacoesRC}
L.~d.~F. Costa, O.~N. Oliveira~Jr, G.~Travieso, F.~A. Rodrigues, P.~R.
  Villas~Boas, L.~Antiqueira, M.~P. Viana, and L.~E. Correa~Rocha.
\newblock Analyzing and modeling real-world phenomena with complex networks: a
  survey of applications.
\newblock {\em Advances in Physics}, 60(3):329--412, 2011.

\bibitem{cnusp}
L.~d.~F. Costa, F.~A. Rodrigues, G.~Travieso, and P.~R. Villas~Boas.
\newblock Characterization of complex networks: A survey of measurements.
\newblock {\em Advances in Physics}, 56(1):167--242, 2007.

\bibitem{dana1999reflectance}
K.~J. Dana, B.~Van~Ginneken, S.~K. Nayar, and J.~J. Koenderink.
\newblock Reflectance and texture of real-world surfaces.
\newblock {\em ACM Transactions on Graphics}, 18(1):1--34, 1999.

\bibitem{geovana2019classification}
G.~V. de~Lima, P.~T. Saito, F.~M. Lopes, and P.~H. Bugatti.
\newblock Classification of texture based on bag-of-visual-words through
  complex networks.
\newblock {\em Expert Systems with Applications}, 2019.

\bibitem{foster1895text}
M.~Foster.
\newblock {\em A Text-book of Physiology}.
\newblock Lea Brothers \& Company, 1895.

\bibitem{gonccalves2012walker}
W.~N. Gon{\c{c}}alves, A.~R. Backes, A.~S. Martinez, and O.~M. Bruno.
\newblock Texture descriptor based on partially self-avoiding deterministic
  walker on networks.
\newblock {\em Expert Systems with Applications}, 39(15):11818--11829, 2012.

\bibitem{gonccalves2016texture}
W.~N. Gon{\c{c}}alves, N.~R. da~Silva, L.~da~Fontoura~Costa, and O.~M. Bruno.
\newblock Texture recognition based on diffusion in networks.
\newblock {\em Information Sciences}, 364:51--71, 2016.

\bibitem{wesley2015dynamic}
W.~N. Gon{\c{c}}alves, B.~B. Machado, and O.~M. Bruno.
\newblock A complex network approach for dynamic texture recognition.
\newblock {\em Neurocomputing}, pages 211--220, 2015.

\bibitem{guo2010completed}
Z.~Guo, L.~Zhang, and D.~Zhang.
\newblock A completed modeling of local binary pattern operator for texture
  classification.
\newblock {\em IEEE Transactions on Image Processing}, 19(6):1657--1663, 2010.

\bibitem{hafner1995efficient}
J.~Hafner, H.~S. Sawhney, W.~Equitz, M.~Flickner, and W.~Niblack.
\newblock Efficient color histogram indexing for quadratic form distance
  functions.
\newblock {\em IEEE transactions on pattern analysis and machine intelligence},
  17(7):729--736, 1995.

\bibitem{haralick}
R.~M. Haralick, K.~Shanmugam, and I.~H. Dinstein.
\newblock Textural features for image classification.
\newblock {\em Systems, Man and Cybernetics, IEEE Transactions on},
  (6):610--621, 1973.

\bibitem{he2016deep}
K.~He, X.~Zhang, S.~Ren, and J.~Sun.
\newblock Deep residual learning for image recognition.
\newblock In {\em Proceedings of the IEEE conference on computer vision and
  pattern recognition}, pages 770--778, 2016.

\bibitem{gabor2005}
M.~A. Hoang, J.-M. Geusebroek, and A.~W. Smeulders.
\newblock Color texture measurement and segmentation.
\newblock {\em Signal processing}, 85(2):265--275, 2005.

\bibitem{huang2002solving}
R.~Huang, Q.~Liu, H.~Lu, and S.~Ma.
\newblock Solving the small sample size problem of lda.
\newblock 2002.

\bibitem{humeau2019texture}
A.~Humeau-Heurtier.
\newblock Texture feature extraction methods: A survey.
\newblock {\em IEEE Access}, 7:8975--9000, 2019.

\bibitem{opponentgabor1998multiscale}
A.~Jain and G.~Healey.
\newblock A multiscale representation including opponent color features for
  texture recognition.
\newblock {\em IEEE Transactions on Image Processing}, 7(1):124--128, 1998.

\bibitem{julesz1962visual}
B.~Julesz.
\newblock Visual pattern discrimination.
\newblock {\em IRE transactions on Information Theory}, 8(2):84--92, 1962.

\bibitem{sajunior2014}
J.~J. d. M.~S. Junior, P.~C. Cortez, and A.~R. Backes.
\newblock Color texture classification using shortest paths in graphs.
\newblock {\em IEEE Transactions on Image Processing}, 23(9):3751--3761, 2014.

\bibitem{keogh2017curse}
E.~Keogh and A.~Mueen.
\newblock Curse of dimensionality.
\newblock In {\em Encyclopedia of machine learning and data mining}, pages
  314--315, Berlin, 2017. Springer.

\bibitem{krizhevsky2012imagenet}
A.~Krizhevsky, I.~Sutskever, and G.~E. Hinton.
\newblock Imagenet classification with deep convolutional neural networks.
\newblock In {\em Advances in neural information processing systems}, pages
  1097--1105, 2012.

\bibitem{lin2013network}
M.~Lin, Q.~Chen, and S.~Yan.
\newblock Network in network.
\newblock {\em arXiv preprint arXiv:1312.4400}, 2013.

\bibitem{liu2019bow}
L.~Liu, J.~Chen, P.~Fieguth, G.~Zhao, R.~Chellappa, and M.~Pietik{\"a}inen.
\newblock From bow to cnn: Two decades of texture representation for texture
  classification.
\newblock {\em International Journal of Computer Vision}, 127(1):74--109, 2019.

\bibitem{maenpaa2004}
T.~M{\"a}enp{\"a}{\"a} and M.~Pietik{\"a}inen.
\newblock Classification with color and texture: jointly or separately?
\newblock {\em Pattern recognition}, 37(8):1629--1640, 2004.

\bibitem{ojala2002outex}
T.~Ojala, T.~Maenpaa, M.~Pietikainen, J.~Viertola, J.~Kyllonen, and
  S.~Huovinen.
\newblock Outex-new framework for empirical evaluation of texture analysis
  algorithms.
\newblock In {\em Pattern Recognition, 2002. Proceedings. 16th International
  Conference on}, volume~1, pages 701--706. IEEE, 2002.

\bibitem{ojala1996comparative}
T.~Ojala, M.~Pietik{\"a}inen, and D.~Harwood.
\newblock A comparative study of texture measures with classification based on
  featured distributions.
\newblock {\em Pattern recognition}, 29(1):51--59, 1996.

\bibitem{ojala2002multiresolution}
T.~Ojala, M.~Pietikainen, and T.~Maenpaa.
\newblock Multiresolution gray-scale and rotation invariant texture
  classification with local binary patterns.
\newblock {\em IEEE Transactions on pattern analysis and machine intelligence},
  24(7):971--987, 2002.

\bibitem{ojansivu2008blur}
V.~Ojansivu and J.~Heikkil{\"a}.
\newblock Blur insensitive texture classification using local phase
  quantization.
\newblock In {\em International conference on image and signal processing},
  pages 236--243. Springer, 2008.

\bibitem{ribas2018fusion}
L.~C. Ribas, J.~J. Junior, L.~F. Scabini, and O.~M. Bruno.
\newblock Fusion of complex networks and randomized neural networks for texture
  analysis.
\newblock {\em arXiv preprint arXiv:1806.09170}, 2018.

\bibitem{LDA}
B.~D. Ripley.
\newblock {\em Pattern recognition and neural networks}.
\newblock Cambridge university press, 1996.

\bibitem{rosenfeld1980multispectral}
A.~Rosenfeld, W.~Cheng-Ye, and A.~Y. Wu.
\newblock Multispectral texture.
\newblock Technical report, MARYLAND UNIV COLLEGE PARK COMPUTER VISION LAB,
  1980.

\bibitem{imagenet}
O.~Russakovsky, J.~Deng, H.~Su, J.~Krause, S.~Satheesh, S.~Ma, Z.~Huang,
  A.~Karpathy, A.~Khosla, M.~Bernstein, et~al.
\newblock Imagenet large scale visual recognition challenge.
\newblock {\em International Journal of Computer Vision}, 115(3):211--252,
  2015.

\bibitem{scabini2019multilayer}
L.~F. Scabini, R.~H. Condori, W.~N. Gonçalves, and O.~M. Bruno.
\newblock Multilayer complex network descriptors for color-texture
  characterization.
\newblock {\em Information Sciences}, 491:30 -- 47, 2019.

\bibitem{scabini2015texture}
L.~F. Scabini, W.~N. Gon{\c{c}}alves, and A.~A. Castro~Jr.
\newblock Texture analysis by bag-of-visual-words of complex networks.
\newblock In {\em Iberoamerican Congress on Pattern Recognition}, pages
  485--492. Springer International Publishing, 2015.

\bibitem{simonyan2014very}
K.~Simonyan and A.~Zisserman.
\newblock Very deep convolutional networks for large-scale image recognition.
\newblock {\em arXiv preprint arXiv:1409.1556}, 2014.

\bibitem{szegedy2015going}
C.~Szegedy, W.~Liu, Y.~Jia, P.~Sermanet, S.~Reed, D.~Anguelov, D.~Erhan,
  V.~Vanhoucke, and A.~Rabinovich.
\newblock Going deeper with convolutions.
\newblock In {\em The IEEE Conference on Computer Vision and Pattern
  Recognition (CVPR)}, June 2015.

\bibitem{szegedy2016rethinking}
C.~Szegedy, V.~Vanhoucke, S.~Ioffe, J.~Shlens, and Z.~Wojna.
\newblock Rethinking the inception architecture for computer vision.
\newblock In {\em The IEEE Conference on Computer Vision and Pattern
  Recognition (CVPR)}, June 2016.

\bibitem{tharwat2017linear}
A.~Tharwat, T.~Gaber, A.~Ibrahim, and A.~E. Hassanien.
\newblock Linear discriminant analysis: a detailed tutorial.
\newblock {\em AI Communications}, 30(2):169--190, 2017.

\bibitem{vandenbroucke2003color}
N.~Vandenbroucke, L.~Macaire, and J.-G. Postaire.
\newblock Color image segmentation by pixel classification in an adapted hybrid
  color space. application to soccer image analysis.
\newblock {\em Computer Vision and Image Understanding}, 90(2):190--216, 2003.

\bibitem{fractal}
M.~Varma and R.~Garg.
\newblock Locally invariant fractal features for statistical texture
  classification.
\newblock In {\em Computer Vision, 2007. ICCV 2007. IEEE 11th International
  Conference on}, pages 1--8. IEEE, 2007.

\bibitem{smallworldCN}
D.~J. Watts and S.~H. Strogatz.
\newblock Collective dynamics of ‘small-world’networks.
\newblock {\em nature}, 393(6684):440--442, 1998.

\bibitem{wichmann2002contributions}
F.~A. Wichmann, L.~T. Sharpe, and K.~R. Gegenfurtner.
\newblock The contributions of color to recognition memory for natural scenes.
\newblock {\em Journal of Experimental Psychology: Learning, Memory, and
  Cognition}, 28(3):509, 2002.

\end{thebibliography}

\end{document}